\documentclass[3p,twocolumn,authoryear]{elsarticle} 

\usepackage[breaklinks=true]{hyperref}

\usepackage{graphicx}
\usepackage[caption=false]{subfig}
\usepackage{adjustbox}
\usepackage{amsmath,amssymb,amsfonts}
\usepackage{algorithmic}
\usepackage{textcomp}
\usepackage{xcolor}
\usepackage{graphics} 
\usepackage{epsfig} 
\usepackage{color}
\usepackage{booktabs} 
\usepackage{siunitx} 
\usepackage{multirow} 
\usepackage{pdflscape} 
\usepackage{afterpage} 
\usepackage[flushleft]{threeparttable} 
\usepackage{makecell}
\usepackage{stfloats}
\usepackage[normalem]{ulem}

\usepackage{import}
\usepackage{flushend}

\usepackage[all]{nowidow}

\DeclareSIUnit\px{px}

\def\BibTeX{{\rm B\kern-.05em{\sc i\kern-.025em b}\kern-.08em
    T\kern-.1667em\lower.7ex\hbox{E}\kern-.125emX}}

\journal{Journal of Field Robotics}

\usepackage{amsmath}
\usepackage{amssymb}
\usepackage{amsopn}
\usepackage{mathtools}

\renewcommand{\vec}[1]{\boldsymbol{\mathbf{#1}}}

\DeclareMathOperator*{\argmin}{arg\,min}

\newcommand{\worldCoordSystem}{\mathit{W}}
\newcommand{\bodyCoordSystem}{\mathit{B}}
\newcommand{\firstGPSCoordSystem}{\mathit{A}}

\newcommand{\coordIndex}[2]{{#1}_{#2}}

\newcommand{\noisyMeasurement}{\hat{\vec{z}}}

\newcommand{\rotation}{\vec{R}}
\newcommand{\rotationCoord}[2]{\rotation_{#2}^{#1}}
\newcommand{\translation}{\vec{t}}
\newcommand{\translationCoord}[2]{\translation_{#2}^{#1}}
\newcommand{\rigidTransform}{\vec{T}}
\newcommand{\rigidTransformCoord}[2]{\rigidTransform_{#2}^{#1}}

\newcommand{\linearVelocity}{\vec{v}}

\newcommand{\systemState}{\vec{x}}
\newcommand{\bias}{\vec{b}}

\renewcommand\cite{\citep} 
\begin{document}

\onecolumn
\begin{center}
This paper has been accepted for publication in \emph{Journal of Field Robotics}.\\
This is the pre-peer reviewed version of the following article:
\\~\\
Cremona J., Civera J., Kofman E. \& Pire T. (2023) GNSS-Stereo-Inertial SLAM for Arable Farming. Journal of Field Robotics, DOI: \nolinkurl{https://doi.org/10.1002/rob.22232} 
\\~\\
which has been published in final form at \url{https://doi.org/10.1002/rob.22232}. This article may be used for non-commercial purposes in accordance with Wiley Terms and Conditions for Use of Self-Archived Versions.

\end{center}

\begin{frontmatter}

\title{GNSS-Stereo-Inertial SLAM for Arable Farming}

\author[add1]{Javier Cremona\corref{mycorrespondingauthor}}
\ead{cremona(at)cifasis-conicet(dot)gov(dot)ar}
\author[add2]{Javier Civera}
\ead{jcivera(at)unizar(dot)es}
\author[add1]{Ernesto Kofman}
\ead{kofman(at)cifasis-conicet(dot)gov(dot)ar}
\author[add1]{Taihú Pire}
\ead{pire(at)cifasis-conicet(dot)gov(dot)ar}
\address[add1]{CIFASIS, French Argentine International Center for Information and Systems Sciences (CONICET-UNR), Rosario, Argentina}
\address[add2]{University of Zaragoza, Zaragoza, Spain}

\cortext[mycorrespondingauthor]{Corresponding author.}

\begin{abstract}
The accelerating pace in the automation of agricultural tasks demands highly accurate and robust localization systems for field robots. Simultaneous Localization and Mapping (SLAM) methods inevitably accumulate drift on exploratory trajectories and primarily rely on place revisiting and loop closing to keep a bounded global localization error. Loop closure techniques are significantly challenging in agricultural fields, as the local visual appearance of different views is very similar and might change easily due to weather effects. A suitable alternative in practice is to employ global sensor positioning systems jointly with the rest of the robot sensors.
In this paper we propose and implement the fusion of GNSS, stereo views and inertial measurements for localization purposes. Specifically, we incorporate, in a tightly-coupled manner, GNSS measurements into the stereo-inertial ORB-SLAM3 pipeline. We thoroughly evaluate our implementation in the sequences of the Rosario dataset \cite{pire2019rosario}, recorded by an autonomous robot in soybean fields, and our own in-house data. Our data includes measurements from a conventional GNSS, rarely included in evaluations of state-of-the-art approaches. We characterize the performance of GNSS-Stereo-inertial SLAM in this application case, reporting pose error reductions between \SI{10}{\percent} and \SI{30}{\percent} compared to visual-inertial and loosely-coupled GNSS-stereo-inertial baselines. In addition to such analysis, we also release the code of our implementation as open source.

\end{abstract}


\end{frontmatter}


\section{Introduction}
\label{sec:introduction}

\begin{figure}[!htbp]
    \centering
    \subfloat[\label{subfig:robot_front}]{\includegraphics[width=0.4\columnwidth]{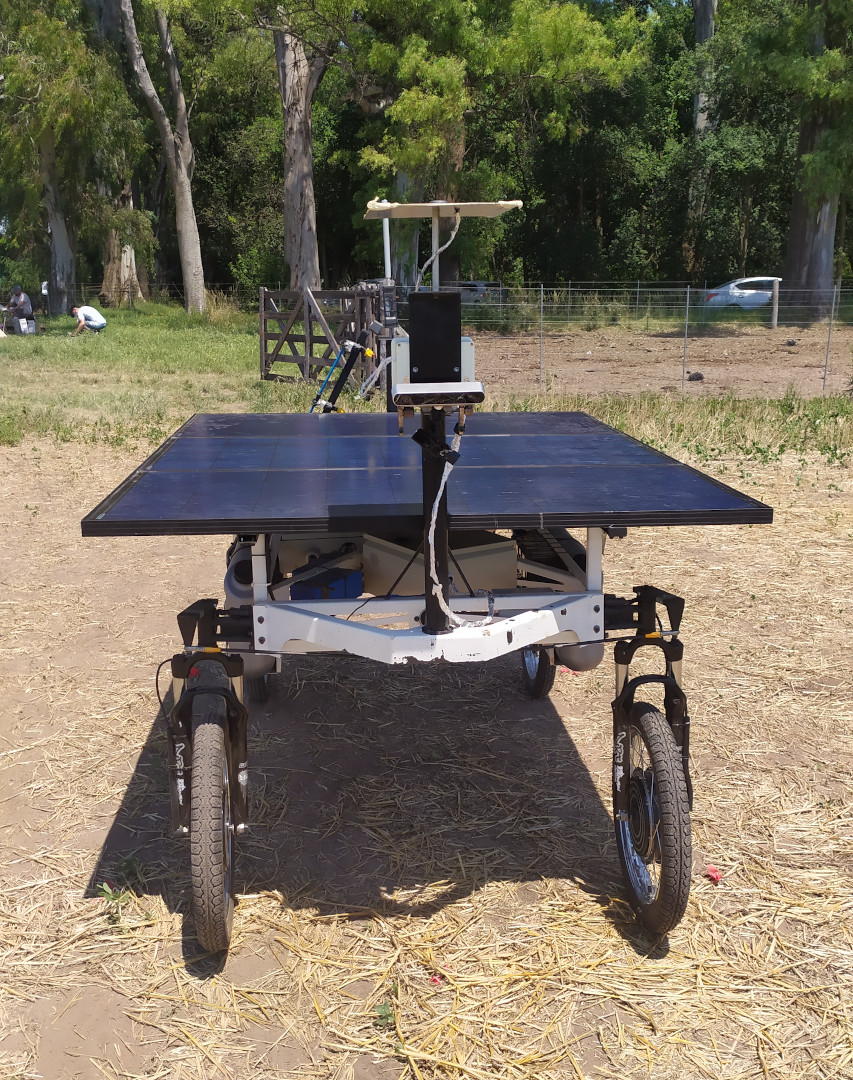}}
    \hspace{0.1em}
    \subfloat[\label{subfig:robot_back}]{\includegraphics[width=0.4\columnwidth]{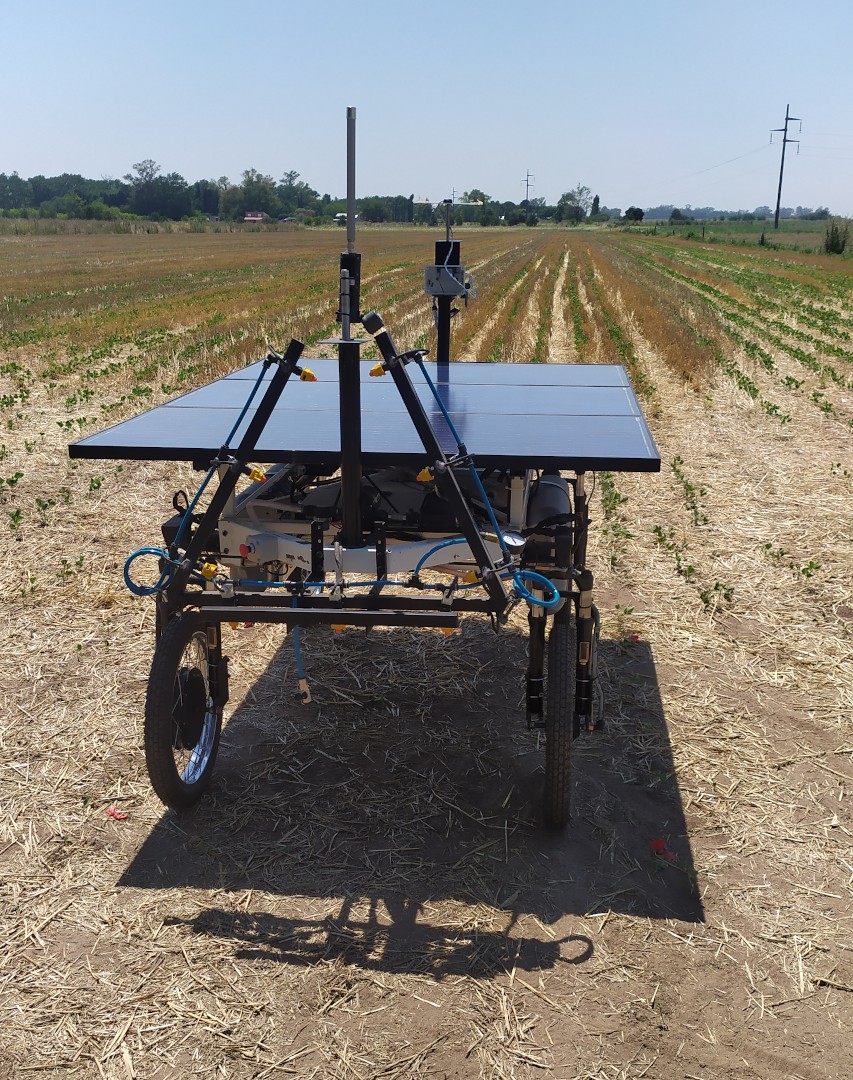}}\\
    \subfloat[\label{subfig:trajectory}]{\includegraphics[width=.9\columnwidth]{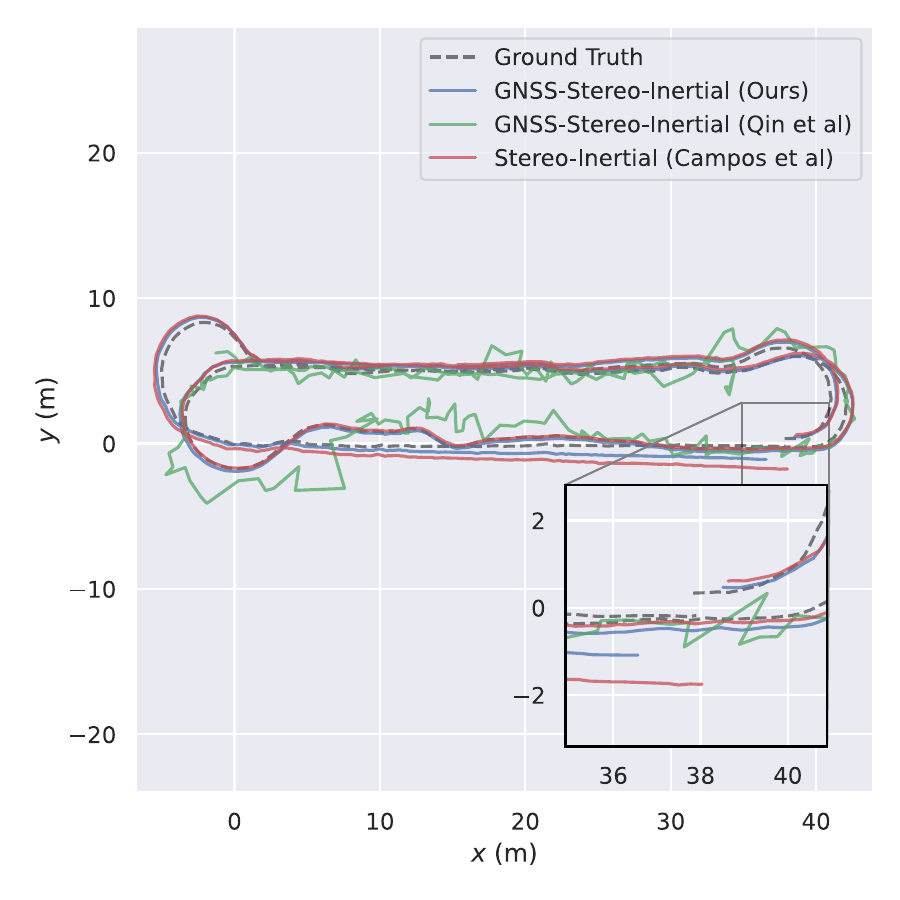}}
    \caption{\protect\subref{subfig:robot_front} and \protect\subref{subfig:robot_back}: Frontal and back views of our field robot and the arable field environment in which we navigate. \protect\subref{subfig:trajectory}: Trajectory estimated by our GNSS-stereo-inertial SLAM framework, along with GNSS-RTK ground truth,  visual-inertial ORB-SLAM3 \cite{campos2021orbslam3} and VINS-Fusion \cite{qin2019general}}
    \label{fig:teaser_image}
\end{figure}

Over the last decades, several agricultural tasks such as sowing, weed detection and removal or harvesting are being progressively automated targeting sustainable and environmentally friendly production. The use of autonomous robots in an agricultural environment has gained relevance, as it enables an efficient use of resources \cite{carelli2013agriculturalrobotics, wouter2014harvesting}. In general, in order to fully automate these and other agricultural tasks, the robot needs to know its pose relative to the environment in which it is navigating. 

A localization system must have a very high degrees of robustness and accuracy for a mobile robot to navigate safely without damaging the environment or itself. For most environments and tasks, a single sensor may not offer a sufficiently reliable robot pose estimate. As a few illustrative examples, GNSS sensors in outdoor environments do not accumulate error (drift) but they present considerable variance in their global position readings and may suffer frequent signal loss. State-of-the-art methods based on visual sensors perform badly if images have insufficient or repetitive textures, which is common in agricultural environments. Lighting can also be a problem if it is insufficient or excessive, and abrupt robot motion can cause image blur that degrades the estimation performance. Finally, interoceptive sensors that measure the internal state of the robot, such as the encoders in the wheel motors or inertial measurement units (IMU), are accurate for short-term motion estimation but drift after a few metres. Summing up, as all sensors have different and complementary advantages and disadvantages, it is essential for field robotics to properly fuse the measurements of multiple sensors to achieve robust and accurate pose estimates. This is particularly relevant to allow the robot to navigate over long periods of time (long-term navigation) and to keep the error bounded locally and globally.

SLAM, standing for Simultaneous Localization and Mapping, stands for the set of methods targeting global localization and mapping from a set of onboard sensors in a mobile agent \cite{cadena2016past}. A large number of visual-inertial SLAM pipelines have been proposed in the last decade \cite{mur2017visual,qin2018vins,campos2021orbslam3}. Many of them demonstrate high accuracy and robustness in indoor and urban environments. However, when it comes to the agricultural environment, they present problems in correctly estimating the pose of the robot. Among others, agricultural environments are challenging for visual navigation due to insufficient and/or repetitive texture and direct sunlight. Adding inertial measurements provides a slight improvement in the estimation. Nevertheless, as shown in \cite{cremona2022evaluation}, state-of-the-art visual-inertial systems accumulate significant errors after navigating a few minutes on arable lands. Robust SLAM systems such as ORB-SLAM3 \cite{campos2021orbslam3} can eliminate drift when revisiting already mapped places, but the so-called loop closing offers a poor performance on agricultural fields due to insufficiently discriminative visual appearances. A reasonable alternative, that we use in this work, is to employ measurements from global positioning sensors such as GNSS to allow the robot to navigate for long periods without accumulating drift.

This paper presents a GNSS-stereo-inertial SLAM implementation that fuses GNSS, visual and inertial measurements using a tightly-coupled approach. Specifically, we extend the state-of-the-art ORB-SLAM3 \cite{campos2021orbslam3} with GNSS factors. The global positioning measurements are incorporated into the mapping thread, so that it performs periodic corrections in the local map and hence also correct the current camera pose in the tracking thread. In this manner, we can achieve drift-less trajectories without depending on the ability of the system to close loops based on visual appearance. We evaluated our implementation on the agricultural dataset known as Rosario Dataset \cite{pire2019rosario} and an additional in-house dataset, which contains data from a wheeled robot in a soybean field (see Figure \ref{subfig:robot_front}-\ref{subfig:robot_back} for a picture of our robot). In both cases, we show how our implementations is able to effectively fuse GNSS readings outperforming the original stereo-inertial ORB-SLAM3. The contribution of the work can be summarized as follows:
\begin{itemize}
    \item Implementation of a GNSS-Stereo-Inertial framework.
    \item Evaluation of our GNSS-Stereo-Inertial framework tightly-coupled fusion in agricultural environments, incorporating real conventional GNSS measurements instead of simulated ones, which are rarely included in evaluations of state-of-the-art approaches.
    \item Public release of our implementation as open-source\footnote{\href{https://github.com/CIFASIS/gnss-stereo-inertial-fusion}{\nolinkurl{https://github.com/CIFASIS/gnss-stereo-inertial-fusion}}} , in order to facilitate its usage, extensions and comparisons and evaluations by the robotics community.
\end{itemize}

The article is organized as follows: Section~\ref{sec:related} discusses related work on multi-modal sensor fusion. In  Section~\ref{sec:method}, we describe the proposed GNSS-Stereo-Inertial framework. In Section~\ref{sec:experiments}, we present and discuss the experimental results of our GNSS-Stereo-Inertial implementation on real data in an agricultural field. Finally, we present our conclusions in Section~\ref{sec:conclusions}.

\section{Related Work}
\label{sec:related}
Sensor fusion methods can be broadly divided into two groups, \emph{loosely-coupled} and \emph{tightly-coupled}. Loosely-coupled methods are those that omit correlations between measurements from different sensors. This simplifies the fusion, as the estimation from each sensor can run separately and the estimates be fused afterwards. Most of these approaches are based on filters, such as the Extended Kalman Filter (EKF), that sequentially updates the system state integrating previous information. This is however suboptimal compared to tightly-coupled methods \cite{strasdat2012why}, which model the correlations between state variables and sensor measurements. In this last case, the measurements from all sensors are jointly integrated in the same optimization problem. As a drawback, tightly-coupled solutions generally have a higher computational cost than loosely-coupled ones. In the rest of the section, we refer the most related works to ours, from the loosely-coupled to the tightly-coupled ones.

\citet{weiss2012versatile} propose an EKF-based estimation method for Micro Air Vehicles (MAV). Its contribution is a modular loosely-coupled method that is capable of fusing visual, inertial and external positioning sensor (such as GPS or a laser telemetry tracking system) information. The results show that the proposed method allows state predictions to be made up to \SI{1}{\kilo\hertz} for MAV control tasks, being robust to low frequency measurements of \SI{1}{\hertz}, delays of up to \SI{500}{\milli\second} in the measurements and noise with standard deviations up to \SI{20}{\centi\meter}. \citet{shen2014multi} presents a similar loosely-coupled approach but using an Unscented Kalman Filter (UKF), in order to better address the non-linearities in the sensor models.  \citet{wei2011intelligent} use stereo cameras to estimate the motion of a ground robot, considering motions only in the horizontal plane, and using an EKF to fuse global GPS measurements in a loosely-coupled manner, reducing the drift. \citet{won2014selective,Won2014gnss} propose a selective integration method for GNSS, visual and inertial measurements to improve localization accuracy under GNSS-challenged environments. The authors introduced a new performance index to recognize poor environments based on the geometrical distribution of the satellites and the local image features. 

\citet{li2019tight} present a multi-state constraint Kalman filter (MSCKF) approach to fuse monocular, inertial and raw GNSS-RTK measurements. The MSCKF makes use of a measurement model that does not require to include the feature landmarks in the state vector of the EKF, improving the robustness and computational complexity of the system. \citet{salehi2017hybrid} use a mixture of tightly-coupled and loosely-coupled techniques for the fusion of visual and GPS measurements. An exhaustive optimization restricted to a temporal window of recent visual measurements is used, while measurements outside the window are marginalized by obtaining estimates of relative motion between poses. This allows to improve computational times, preventing the computational complexity to scale. \citet{yu2019gpsaided} present a GPS-assisted visual-inertial estimation framework for omnidirectional platforms. It extends VINS-MONO \cite{qin2018vins} to support multiple cameras, fuses visual and inertial information in a tightly-coupled manner, combined with a loosely-coupled approach to incorporate the measurements provided by GPS. Later, the same authors present GVINS \cite{cao2021gvins}, a framework based on non-linear optimization. GVINS tightly fuses GNSS raw measurements with visual and inertial information for state estimation. The GNSS pseudorange and Doppler shift measurements are modelled under a probabilistic factor graph framework along with visual and inertial constraints. The same approach is applied in \cite{liu2021optimization}.

\citet{lynen2013robust} present Multi-Sensor Fusion (MSF), a modular sensor fusion system based on an EKF filter where inertial information is used at the prediction step. The information coming from the different sensors is modeled in a general manner as relative and/or absolute pose estimates, thus allowing to fuse measurements coming from a large number of sensors using a loosely-coupled approach. The work places particular emphasis on modelling the temporal arrival of the measurements by applying a technique known as Stochastic Cloning able to address asynchronous sensor fusion. \citet{mascaro2018gomsf} present the Graph-Optimization based Multi-Sensor Fusion (GOMSF) framework which solves the fusion of pose estimates in different coordinate systems. Visual-inertial estimates from the MSF in local coordinates are merged with measurements in global coordinates from a GPS. 

\citet{lee2020intermittent} present a GPS-VIO system that fuses visual-inertial data with intermittent GPS measurements is presented. The authors proposed a GPS-IMU online calibration approach for the time offset and extrinsics estimation. In \cite{boche2022dropout} a tightly coupled visual-inertial-GPS system is presented. The system is based on OKVIS2 \cite{leutenegger2022okvis2}. In the work a new global reference frame initialisation has been introduced that incorporates measurement uncertainties to decide whether the extrinsic transformation between the global and visual-inertial reference frame becomes observable.

In contrast to the previously mentioned works, this paper presents a tightly-coupled GNSS-stereo-inertial SLAM to tackle localization in agricultural environments. The proposed framework extends the Visual-Inertial SLAM system ORB-SLAM3 \cite{campos2021orbslam3} with GNSS measurements. We built on top of ORB-SLAM3 since it has a fair performance in agricultural environments \cite{cremona2022evaluation}. Our implementation is publicly released as open source to facilitate its use, extension and reproduction of the results by the robotics community.


\section{Proposed GNSS-Stereo-Inertial Framework}
\label{sec:method}
This section presents the technical aspects of our implementation. Firstly, we introduce the notation and conventions adopted that are necessary to fully detail the model of our GNSS factor. Later, we briefly introduce ORB-SLAM3 \cite{campos2021orbslam3}, the state-of-the-art framework Visual-Inertial SLAM that we use in our method. We refer the reader to the original ORB-SLAM3 publication for the full details on such framework. Finally, we detail the formulation of our GNSS factor.

\subsection{Notation}
\begin{figure}[!t]
    \centering
    \includegraphics[width=\linewidth]{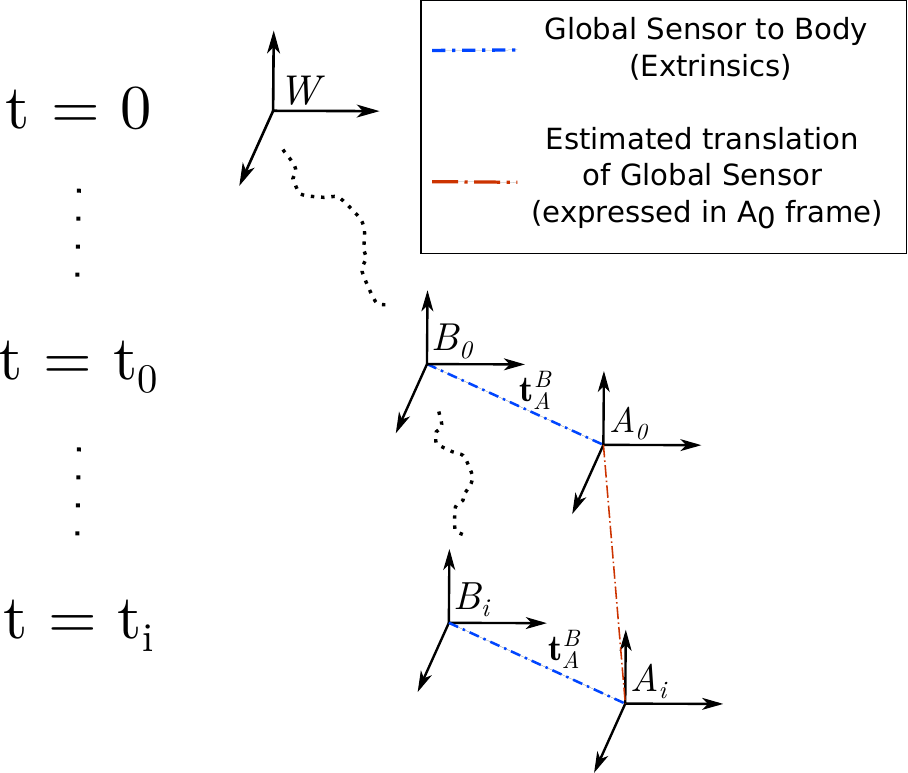}
    \caption{Reference Frames used in this work. The position of the GNSS antenna in the body frame is represented with a translation and shown with a blue line, and can be obtained from the calibration of the system. The red line represents the estimated translation of the GNSS antenna from time $t_0$ to time $t_i$. This 3D vector is compared with the GNSS measurements in the GNSS error residual $\mathbf{r}_{\mathcal{G}_{i}}$. Both vectors are expressed in ${\coordIndex{\firstGPSCoordSystem}{0}}$ frame.}
    \label{fig:frames}
\end{figure}

Figure~\ref{fig:frames} shows the coordinate frames used in this work. $\worldCoordSystem$ represents the world frame and $\bodyCoordSystem$ represents the body frame, that we place in the IMU sensor. $\vec{a}^{S}$ represents the coordinates of a geometry entity $\vec{a}$ with respect to the reference frame $S$. $\rotationCoord{\worldCoordSystem}{\bodyCoordSystem} \in SO(3)$ refers to the rotation of $\bodyCoordSystem$ with respect to $\worldCoordSystem$, and $\translationCoord{\worldCoordSystem}{\bodyCoordSystem} \in \mathbb{R}^3$ represents the translation of the reference frame $\bodyCoordSystem$ expressed in the frame $\worldCoordSystem$. The rigid transformation formed by the rotation $\rotationCoord{\worldCoordSystem}{\bodyCoordSystem}$ and the translation $\translationCoord{\worldCoordSystem}{\bodyCoordSystem}$ is denoted as $\rigidTransformCoord{\worldCoordSystem}{\bodyCoordSystem} \in SE(3)$, and transforms points in homogeneous coordinates from the reference frame ${\bodyCoordSystem}$ to the reference frame ${\worldCoordSystem}$. For global positioning measurements, $\translationCoord{\bodyCoordSystem}{\firstGPSCoordSystem} \in \mathbb{R}^3$ is the position of the GNSS antenna in the body frame, and is assumed to be known from a calibration stage. All GNSS measurements are transformed to the local Cartesian frame that we denote as ${\coordIndex{\firstGPSCoordSystem}{0}}$. We detail below how we choose such reference frame.

\subsection{ORB-SLAM3}
ORB-SLAM3 is a state-of-the-art visual-inertial SLAM framework evolved from ORB-SLAM2 \cite{mur2017orb} and ORB-SLAM-VI \cite{mur2017visual}. With respect to ORB-SLAM-VI, ORB-SLAM3 proposes a substantially more robust inertial initialization based on maximum-a-posteriori estimates. As it is common in current SLAM systems, the processing is split into multiple threads to exploits multi-core architectures. Specifically, ORB-SLAM3 implements a tracking thread, a local mapping thread and a loop closure and map merging thread. The tracking thread estimates the pose of the current frame by minimizing the reprojection error and incorporating IMU constraints into the optimization by pre-integration \cite{forster2017onmanifold}. It also contains the heuristics for deciding whether a frame becomes a keyframe. The mapping thread main task is a visual-inertial bundle adjustment on a sliding window of keyframes, although it also performs auxiliary map management tasks such as point and keyframe culling. Finally, the loop closure and map merging thread ensures the global consistency of large maps by recognizing revisited places and correcting the drift, and joining separate maps if a common overlap is detected.

From the results in \cite{cremona2022evaluation}, ORB-SLAM3 presents an acceptable accuracy in arable lands for short camera trajectories, but long-term navigation is still challenging. The authors propose a novel loop closure algorithm to correct the drift. However, even with such improvement, loop closure keeps being challenging due to the similarity in appearance of the local visual features. As a result, visual SLAM systems may accumulate drift when loop closures are not detected or the estimation may be corrupted by false loop detections.

\subsection{GNSS-Stereo-Inertial Fusion}
\begin{figure}[!t]
    \centering
    \includegraphics[width=\linewidth]{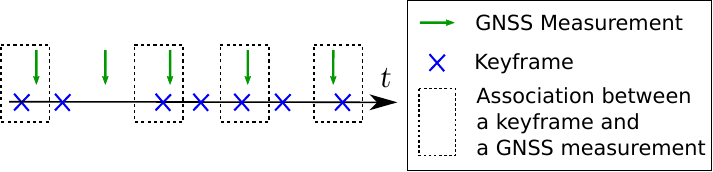}
    \caption{Temporal association between keyframes and GNSS measurements. 
    GNSS measurements are discarded if they are further than specific temporal threshold from any keyframe.}
    \label{fig:association}
\end{figure}

In this work, we formulate a tightly-coupled approach for fusing visual, inertial and GNSS data. Firstly, GNSS measurements are associated to the timestamp of a keyframe according to their temporal proximity. If there is a keyframe with a temporal difference under a specific threshold, the GNSS constraint is set to this keyframe. GNSS readings that are not close in time to any keyframe are discarded (see an illustration of this approach in Figure~\ref{fig:association}). While this is an approximation, we found that, given the high variance of conventional GNSS, a sufficiently small threshold and appropriate keyframe management policy makes its effect negligible. 

The first GNSS reading that is associated with a keyframe determines the position of $\coordIndex{\firstGPSCoordSystem}{0}$, the Cartesian frame for our global position measurements (see Figure~\ref{fig:frames}). We choose $\coordIndex{\firstGPSCoordSystem}{0}$ as a East-North-Up (ENU) local Cartesian frame. The subsequent GNSS measurements are transformed to be expressed in $\coordIndex{\firstGPSCoordSystem}{0}$, and we refer to them as $\noisyMeasurement_{i}$, where $t_i$ is the timestamp of the corresponding keyframe. This is done once the IMU is initialized. If the map is reset, the process of selecting $\coordIndex{\firstGPSCoordSystem}{0}$ is repeated. 

\begin{figure}[!tb]
    \centering
    \includegraphics[width=\columnwidth]{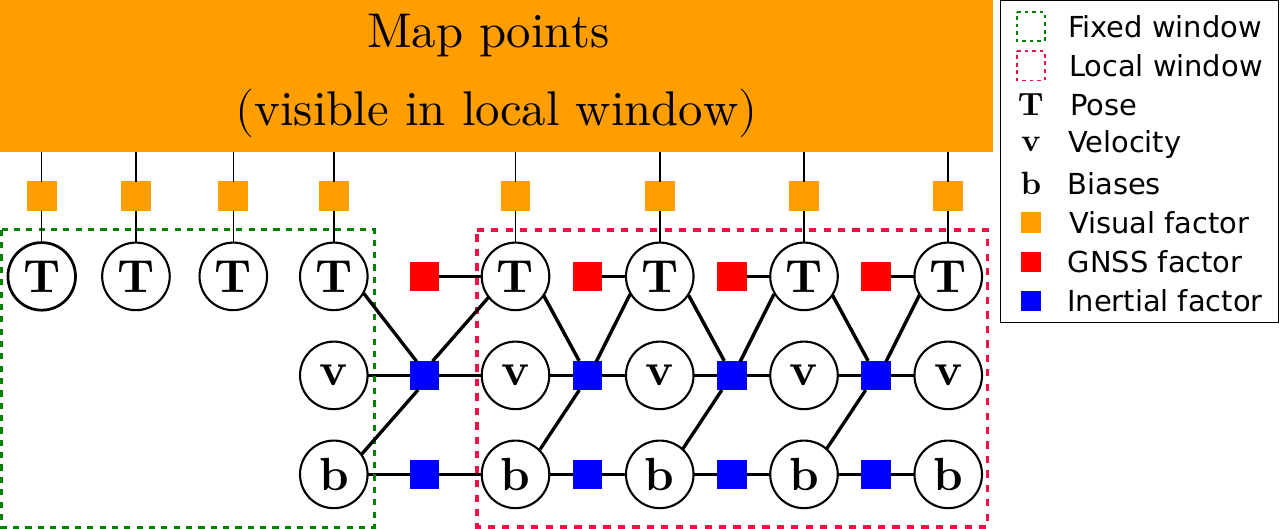}
    \caption{Factor Graph corresponding to the Local Bundle Adjustment of our GNSS-Stereo-Inertial SLAM. In comparison to ORB-SLAM3, a GNSS factor (in red) is added to the cost function. The Local Window is composed by the $N$ last keyframes.}
    \label{fig:graph}
\end{figure}

Our GNSS-Stereo-Inertial fusion is done in the local bundle adjustment of a sliding window of keyframes and 3D points observed from them. Figure~\ref{fig:graph} shows the factor graph corresponding to such optimization. The state variables to optimize are $\mathcal{X} = \{\mathcal{X}_{B}, \mathcal{L}\}$, where $\mathcal{X}_{B} = [\systemState_{1},\dots,\systemState_{i},\dots,\systemState_{N}]$ is the set of sensor states for a window covering the last $N$ keyframes and $\mathcal{L}= [\mathbf{y}_{1},\dots,\mathbf{y}_{j},\dots,\mathbf{y}_{M}]$ is the set of landmarks states that were measured during those last $N$ keyframes. The sensor state $\systemState_{i}$ at the time instant $i$ is
\begin{equation}
    \systemState_{i} = [\rigidTransformCoord{\worldCoordSystem}{\coordIndex{\bodyCoordSystem}{i}}, \coordIndex{\linearVelocity}{i}^\top, \bias_{a_i}^\top, \bias_{g_i}^\top],
\end{equation}
which contains the sensor rigid transformation with respect to the world frame $\rigidTransformCoord{\worldCoordSystem}{\coordIndex{\bodyCoordSystem}{i}} \in SO(3)$, its local velocity $\coordIndex{\linearVelocity}{i} \in \mathbb{R}^3$ and the accelerometer and gyroscope bias $\bias_{a_i} \in \mathbb{R}^3$ and $\bias_{g_i} \in \mathbb{R}^3$. Landmarks are represented by their Euclidean coordinates in the world frame, i.e., $\mathbf{y}_{j} = [X^W, Y^W, Z^W]^\top \in \mathbb{R}^3$

In comparison to ORB-SLAM3, a GNSS error term is added to the cost function. Note that, as shown in Figure~\ref{fig:association}, some keyframes may not have an associated GNSS measurement. 
Then, our GNSS-Stereo-Inertial mapping optimization can be stated as follows
\begin{equation}
\begin{split}
    \hat{\mathcal{X}} = \argmin_{\mathcal{X}} \left( \sum_{i=1}^N \lVert \mathbf{r}_{\mathcal{I}_{i-1,i}}  \rVert^2_{\Sigma_{\mathcal{I}_{i-1,i}}^{-1}} +\right. \\ 
    \left. + \sum_{j=1}^M \sum_{i \in \mathcal{K}_j} \rho \left( \lVert \mathbf{r}_{\mathcal{V}_{ij}} \rVert_{\Sigma_{{V}_{ij}}^{-1}} +\right) \right. \\
    \left. + \sum_{i \in \mathcal{N}^*} \rho \left( \lVert \mathbf{r}_{\mathcal{G}_{i}} \rVert_{\Sigma_{\mathcal{G}_{i}}^{-1}} +\right) \right),
\end{split}
\end{equation}
where $\mathcal{N}^*$ is the set of keyframes that have an associated GNSS measurement. The three addends correspond, respectively, to the inertial, visual and GNSS constraints. For completitude we will detail the three of them, although the first two are used exactly as proposed in ORB-SLAM3 and the third one is our novel contribution.

The inertial residual is defined as follows
\begin{equation}
\mathbf{r}_{\mathcal{I}_{i-1,i}} = [\mathbf{r}_{\Delta \mathbf{R}_{i-1,i}}^\top, \mathbf{r}_{\Delta \mathbf{v}_{i-1,i}}^\top, \mathbf{r}_{\Delta \mathbf{p}_{i-1,i}}^\top]^{\top},
\end{equation}
where $\mathbf{r}_{\Delta \mathbf{R}_{i-1,i}}$, $\mathbf{r}_{\Delta \mathbf{v}_{i-1,i}}$ and $\mathbf{r}_{\Delta \mathbf{p}_{i-1,i}}$ correspond to orientation, velocity and position residuals that have the following form
\begin{equation}
\footnotesize
\begin{aligned}
\mathbf{r}_{\Delta \mathbf{R}_{i-1,i}} &= \log \left( \Delta \mathbf{R}_{i-1,i}^\top \mathbf{R}_{i-1}^\top \mathbf{R}_{i} \right) \\
\mathbf{r}_{\Delta \mathbf{v}_{i-1,i}} &= \mathbf{R}_{i}^\top \left( \mathbf{v}_{i} - \mathbf{v}_{i-1} - \mathbf{g}\Delta t_{i-1,i} \right) - \Delta \mathbf{v}_{i-1,i} \\
\mathbf{r}_{\Delta \mathbf{p}_{i-1,i}} &= \mathbf{R}_{i}^\top \left( \mathbf{p}_{i} - \mathbf{p}_{i-1} - \mathbf{v}_{i} \Delta t_{i-1,i} - \frac{1}{2}\mathbf{g}\Delta t_{i-1,i}^2 \right) - \\ & \ \ \ - \Delta \mathbf{p}_{i-1,i}.
\end{aligned}
\end{equation}
The terms denoted as $\Delta \mathbf{R}_{i-1,i}$, $\Delta \mathbf{v}_{i-1,i}$ and $\Delta \mathbf{p}_{i-1,i} $ come from the preintegration of the IMU readings between the time instants $i-1$ and $i$, and are computed together with their on-manifold covariance $\Sigma_{\mathcal{I}_{i-1,i}}$ according to \cite{forster2017onmanifold}. $\mathbf{g}$ stands for the gravity direction, which is set at the system bootstrapping.

The visual residual $\mathbf{r}_{\Delta \mathbf{v}_{i-1,i}}$ is
\begin{equation}
    \mathbf{r}_{\mathcal{V}_{ij}} = \mathbf{u}_{ij} - \pi\left( \rigidTransformCoord{C}{\bodyCoordSystem} \rigidTransformCoord{\worldCoordSystem-1}{\bodyCoordSystem} \tilde{\mathbf{y}}_{j} \right),
\end{equation}
where $ \tilde{\mathbf{y}}_{j}$ stands for the homogeneous representation of the $j^{th}$ landmark, $\pi(\cdot)$ for the pinhole projection model of a 3D point in homogeneous coordinates in a stereo image, and $\mathbf{u}_{ij}$ the measured image coordinates of the $j^{th}$ landmark in the $i^{th}$ stereo keyframe. The visual covariance of image landmarks $\Sigma_{\mathcal{V}_{ij}}$ is set to the standard 1-pixel standard deviation isotropic Gaussian.

Finally, the GNSS error residual is
\begin{equation}
\footnotesize
\mathbf{r}_{\mathcal{G}_{i}} = \noisyMeasurement_{i} -  \rotationCoord{\coordIndex{\firstGPSCoordSystem}{0}}{\worldCoordSystem}\left(\rotationCoord{\worldCoordSystem}{\coordIndex{\bodyCoordSystem}{i}}\translationCoord{\bodyCoordSystem}{\firstGPSCoordSystem} + \translationCoord{\worldCoordSystem}{\coordIndex{\bodyCoordSystem}{i}} - \left(\rotationCoord{\worldCoordSystem}{\coordIndex{\bodyCoordSystem}{0}}\translationCoord{\bodyCoordSystem}{\firstGPSCoordSystem} + \translationCoord{\worldCoordSystem}{\coordIndex{\bodyCoordSystem}{0}} \right)\right).
\end{equation}
The second term represents the translation vector of the global sensor (in this case, the GNSS antenna) at time instant $i$ in the reference frame $\coordIndex{\firstGPSCoordSystem}{0}$, as can be seen in Figure~\ref{fig:frames}. $\rotationCoord{\worldCoordSystem}{\coordIndex{\bodyCoordSystem}{0}}$ and $\translationCoord{\worldCoordSystem}{\coordIndex{\bodyCoordSystem}{0}}$, which are the relative rotation and translation between the body and the world frame at time $t_0$, are kept constant during the optimization. $\rotationCoord{\coordIndex{\firstGPSCoordSystem}{0}}{\worldCoordSystem}$ is computed by aligning  the first 20 GNSS measurements  with the poses estimated by ORB-SLAM3 in the same time period using Umeyama's method \cite{umeyama1991least}. After estimating this rotation, it is kept fixed during the whole optimization process. The covariance matrix $\Sigma_{\mathcal{G}_{i}}$ is set from the specifications sheet of our GNSS device in each Cartesian axis
\begin{equation}
\Sigma_{\mathcal{G}_{i}} = \begin{bmatrix}
\sigma_x^{2} & 0 & 0\\
0 & \sigma_y^{2} & 0 \\
0 & 0 & \sigma_z^{2}
\end{bmatrix}.
\end{equation}
This covariance matrix is defined relative to a tangential plane through the GNSS reported position. The values are expressed in ENU frame. Finally, the Jacobian with respect to the pose error state is defined as
\begin{equation}
\frac{\partial \mathbf{r}_{\mathcal{G}_{i}}}{\partial \delta \rigidTransformCoord{\worldCoordSystem}{\coordIndex{\bodyCoordSystem}{i}}} = \left[ \rotationCoord{\coordIndex{\firstGPSCoordSystem}{0}}{\worldCoordSystem}  \rotationCoord{\worldCoordSystem}{\coordIndex{\bodyCoordSystem}{i}} \left[ \translationCoord{\bodyCoordSystem}{\firstGPSCoordSystem}\right]^{\times} \quad -\rotationCoord{\coordIndex{\firstGPSCoordSystem}{0}}{\worldCoordSystem}  \right],
\end{equation}

where $\delta$ indicates that the derivative is computed with respect to a right perturbation in the pose.

\section{Experimental Evaluation}
\label{sec:experiments}
This section shows the experimental results of the implementation proposed in Section~\ref{sec:method}. The framework is evaluated on the Rosario Dataset \cite{pire2019rosario}, a set of agricultural data captured by a weed removal robot. Later, an evaluation of the system in a soybean field is presented, using the same weed removal robot. The difference between the latter test and the evaluation on the Rosario Dataset is that new sensors are available, including measurements from a conventional GNSS.

\subsection{Rosario Dataset}
\label{sec:experiments_rosario_dataset}
The Rosario Dataset \cite{pire2019rosario} is a set of data captured by the sensors of a weed removal robot developed by the CIFASIS institute (CONICET-UNR) in Rosario, Argentina. It is composed of six sequences captured in a soybean field. The sequences contain stereo images of $672\times\SI{376}{\px}$ captured at \SI{15}{\hertz}, measurements from an IMU with a frequency of \SI{142}{\hertz} including gyroscope and accelerometer, wheel odometry obtained at \SI{10}{\hertz} and GNSS-RTK measurements at \SI{5}{\hertz}. The GNSS-RTK data are used as positional ground‐truth.

Since the Rosario Dataset does not have conventional GNSS measurements, we simulate noisy GNSS measurements by corrupting the ground-truth with zero-mean Gaussian noise, as in \cite{cioffi2020tightly}. We use isotropic Gaussian noise $\mathbf{n}_{p}\sim\mathcal{N}\left(\mathbf{0},\sigma_{p}^{2}\cdot\mathbf{I}\right)$, with a standard deviation $\sigma_{p} = \SI{0.5}{\meter}$. We selected this value from observing the covariance of the conventional GNSS used in the experiments in Section~\ref{sec:experiments_conventional_gps}. 

\begin{table}[!tp]
    \centering
    \caption{Mean and standard deviation (between parentheses) of the Absolute Trajectory Error (ATE) [\si{\metre}] for stereo-inertial ORB-SLAM3 \cite{campos2021orbslam3}, a loosely-coupled GNSS-stereo-inertial system \cite{qin2019general} and our tightly-coupled GNSS-stereo-inertial framework in the six sequences of the Rosario Dataset. Best results are in \bf{bold}.}
    
    \resizebox{\linewidth}{!} {
        \begin{tabular}{cccc}
        \hline
                    Sequence & Stereo-Inertial & GNSS-Stereo-Inertial & GNSS-Stereo-Inertial \\
                    & \cite{campos2021orbslam3} & \cite{qin2019general} & (Ours) \\
                    \hline
        01 &  0.90 (0.34)                  &
        1.44 (2.06) &
        \bf{0.86 (0.26)}                \\
        02 &  1.33 (0.75)                  &        \bf{0.90 (0.40)} &
        0.94 (0.56)                 \\
        03 &  1.12 (0.65)                  & 1.34 (1.91) & \bf{0.99 (0.56)}                \\
        04 & 1.09 (0.65)                   & 1.42 (1.20) &
        \bf{1.04 (0.60)}                \\
        05 &  0.89 (0.55)                  & 1.43 (1.91) & \bf{0.76 (0.38)}                \\
        06 & 2.48 (1.40)                   & 1.81 (0.87) &  \bf{1.23 (0.70)}     \\ \hline
        \end{tabular}
    }
    \label{tab:rosario_results}
\end{table}

\begin{figure*}[!htp]
  \centering
  \subfloat[Sequence 01\label{trajectories_sequence_01}]{\includegraphics[width=0.42\textwidth]{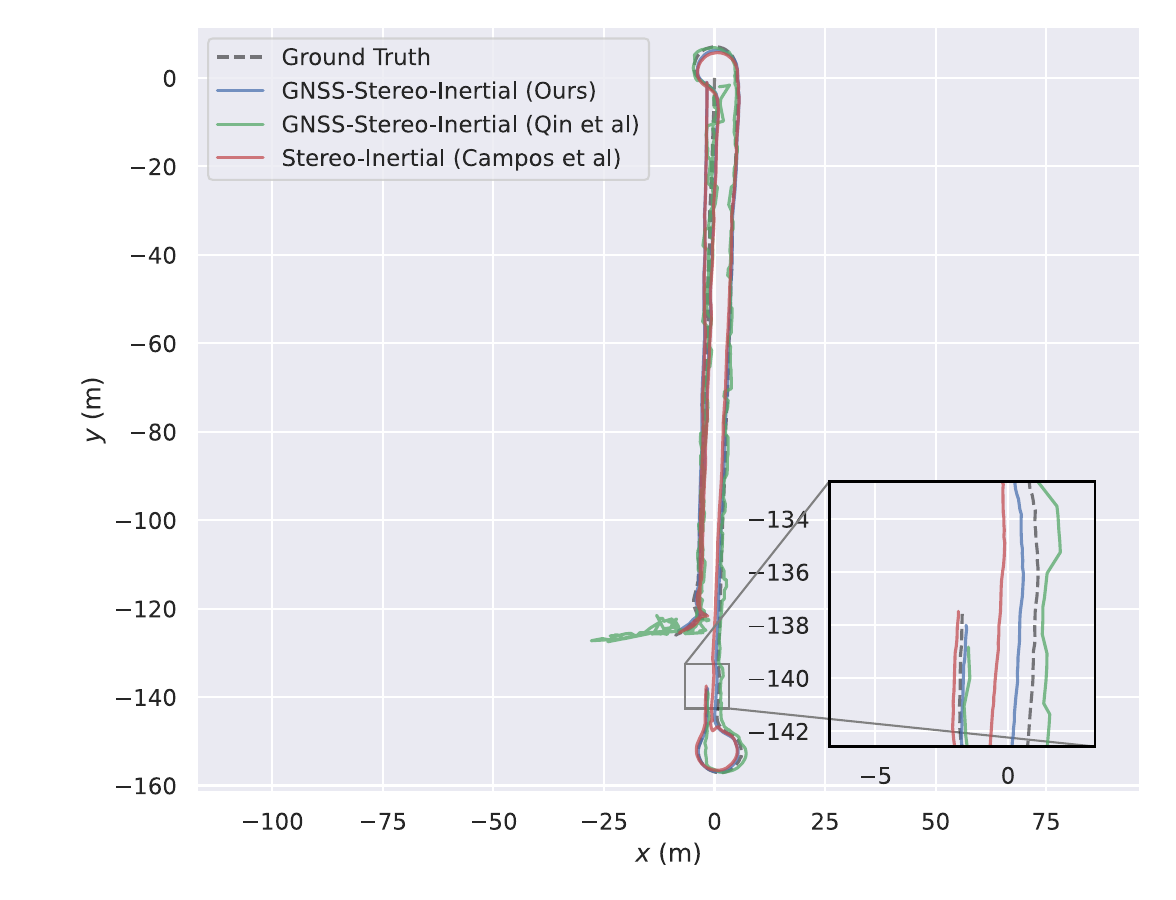}}
  \hspace{1cm}
  \subfloat[Sequence 02\label{trajectories_sequence_02}]{\includegraphics[width=0.42\textwidth]{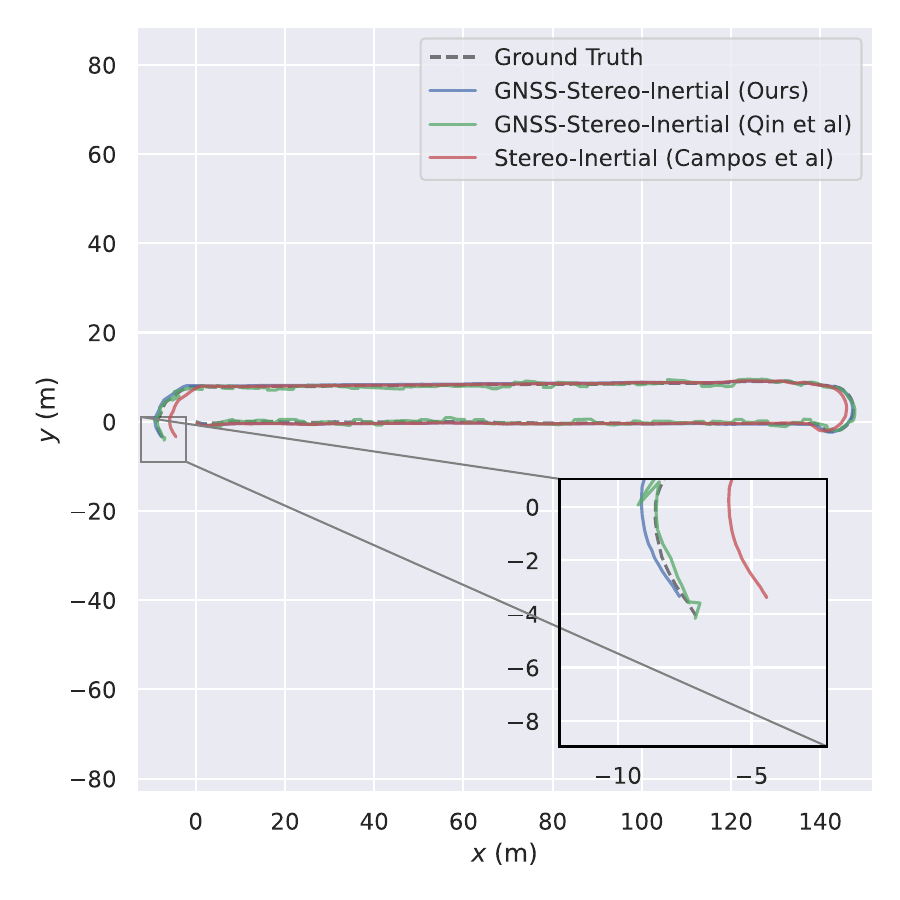}\label{trajectories_sequence_02_int}}\\
  \subfloat[Sequence 03\label{trajectories_sequence_03}]{\includegraphics[width=0.42\textwidth]{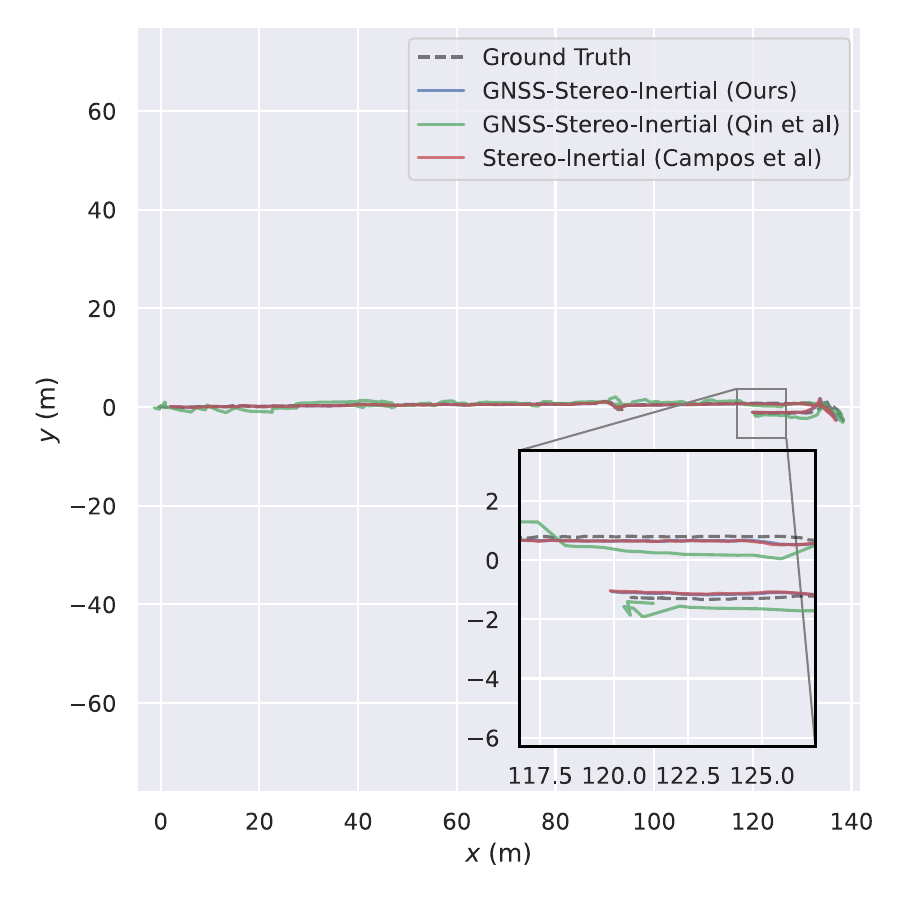}}
  \hspace{1cm}
  \subfloat[Sequence 04\label{trajectories_sequence_04}]{\includegraphics[width=0.42\textwidth]{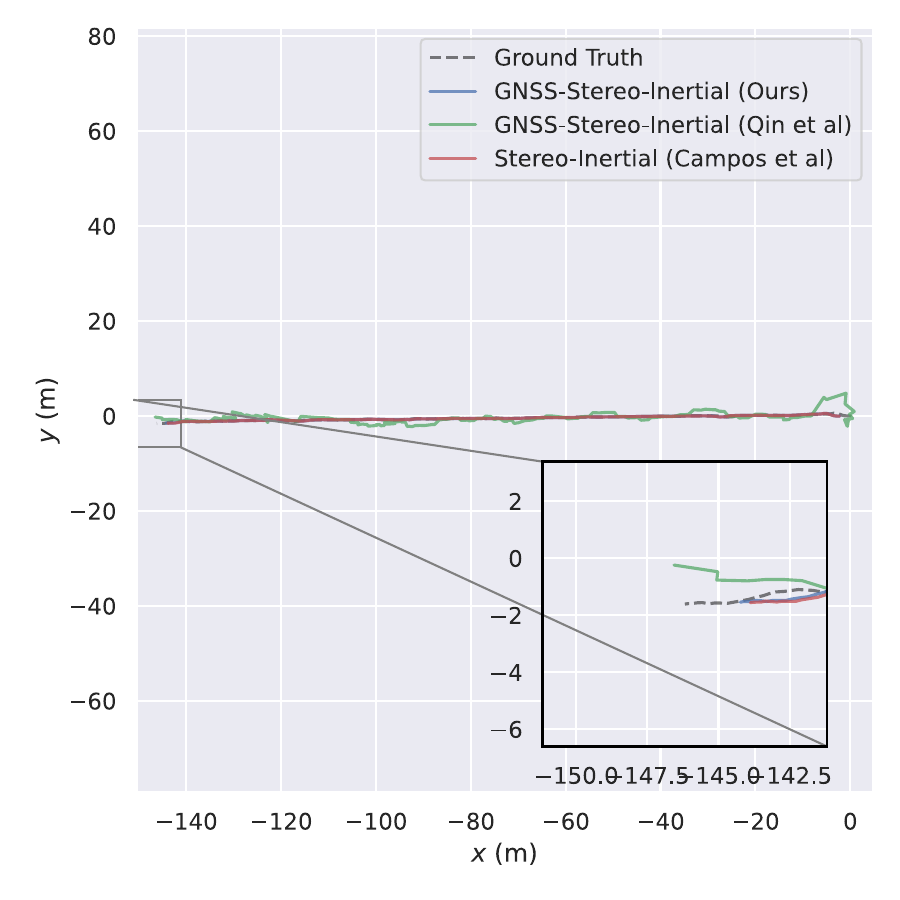}}\\
  \subfloat[Sequence 05\label{trajectories_sequence_05}]{\includegraphics[width=0.42\textwidth]{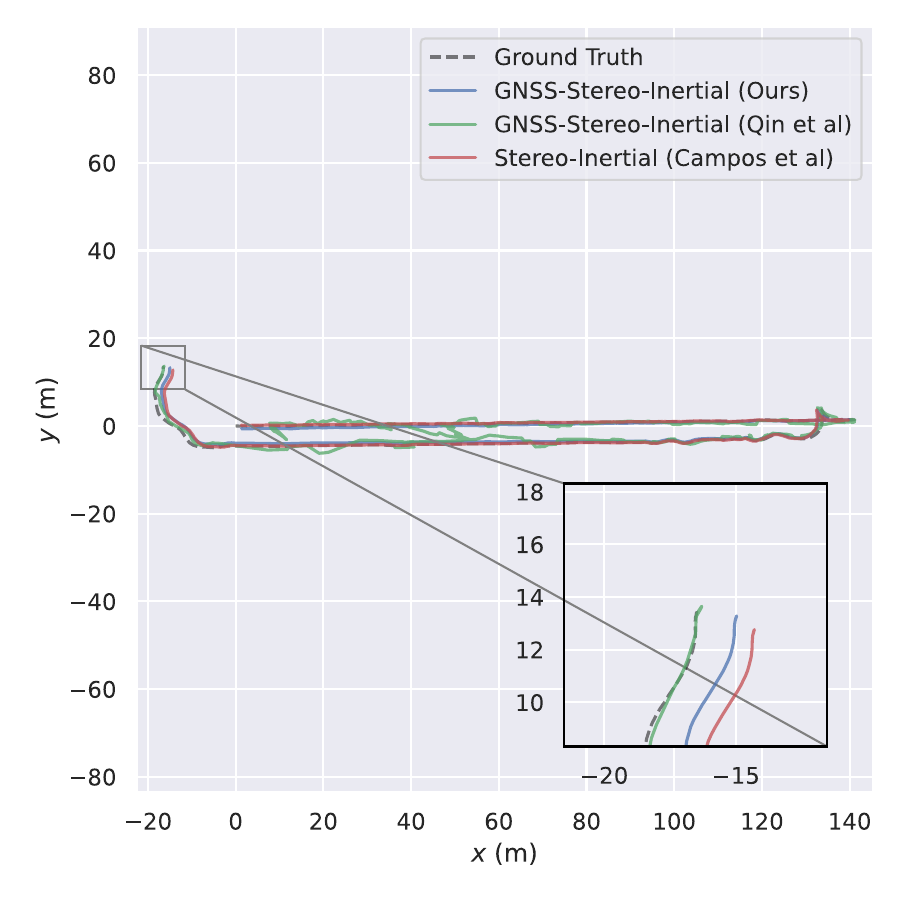}}
  \hspace{1cm}
  \subfloat[Sequence 06\label{trajectories_sequence_06}]{\includegraphics[width=0.42\textwidth]{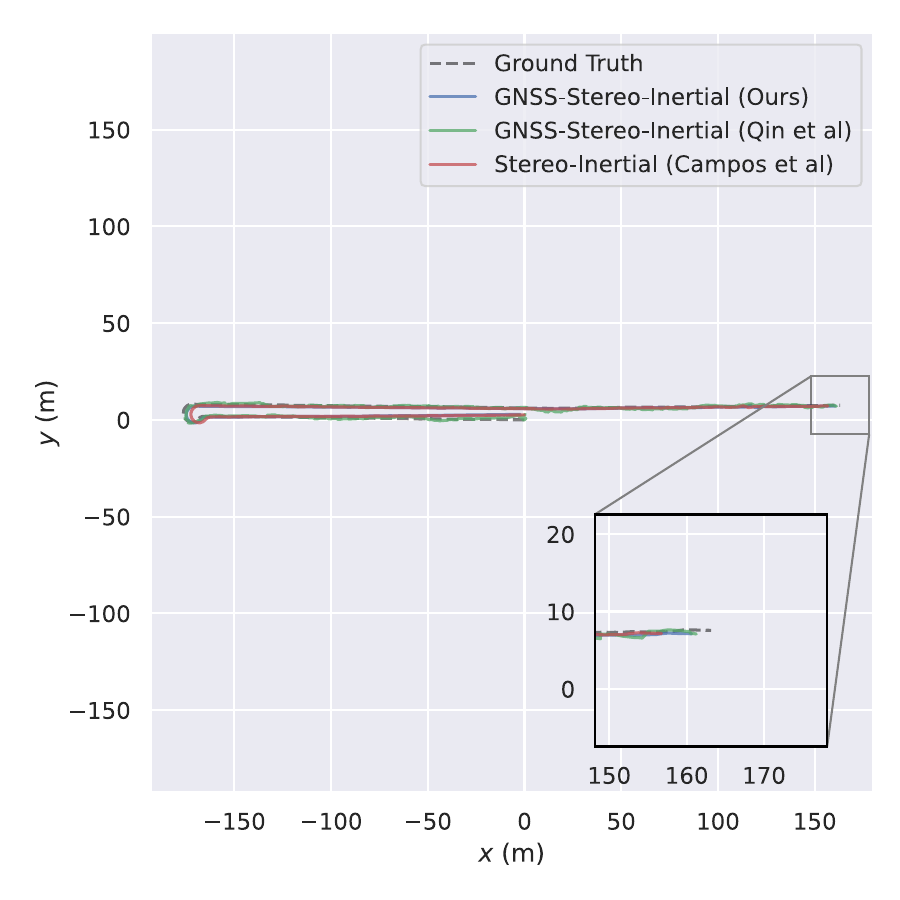}}
  \caption{Results from Stereo-Inertial ORB-SLAM3 \cite{campos2021orbslam3}, a loosely-coupled GNSS-Stereo-Inertial system \cite{qin2019general} and our tightly-coupled GNSS-Stereo-Inertial system on the Rosario Dataset.}
  \label{fig:trajectories_rosario}
\end{figure*}

We compared our GNSS-Stereo-Inertial implementation against Stereo-Inertial ORB-SLAM3 and a loosely-coupled GNSS-Stereo-Inertial system known as VINS-Fusion \cite{qin2019general}. VINS-Fusion was chosen because it is a state-of-the-art system that takes as input the same GNSS measurements as our system, i.e. latitude, longitude and altitude. Each system was run five times in each of the Rosario sequences, and Table~\ref{tab:rosario_results} presents the lowest ATE error of the five executions for each framework. ATE
has been computed after the estimated trajectories
were aligned with the ground-truth GNSS readings using Umeyama's method \cite{umeyama1991least}. The corresponding trajectories are presented in Figure~\ref{fig:trajectories_rosario}. 

\subsection{Data with conventional GNSS in soybean fields}
\label{sec:experiments_conventional_gps}
\begin{figure}[t]
    \centering
    \includegraphics[width=\columnwidth]{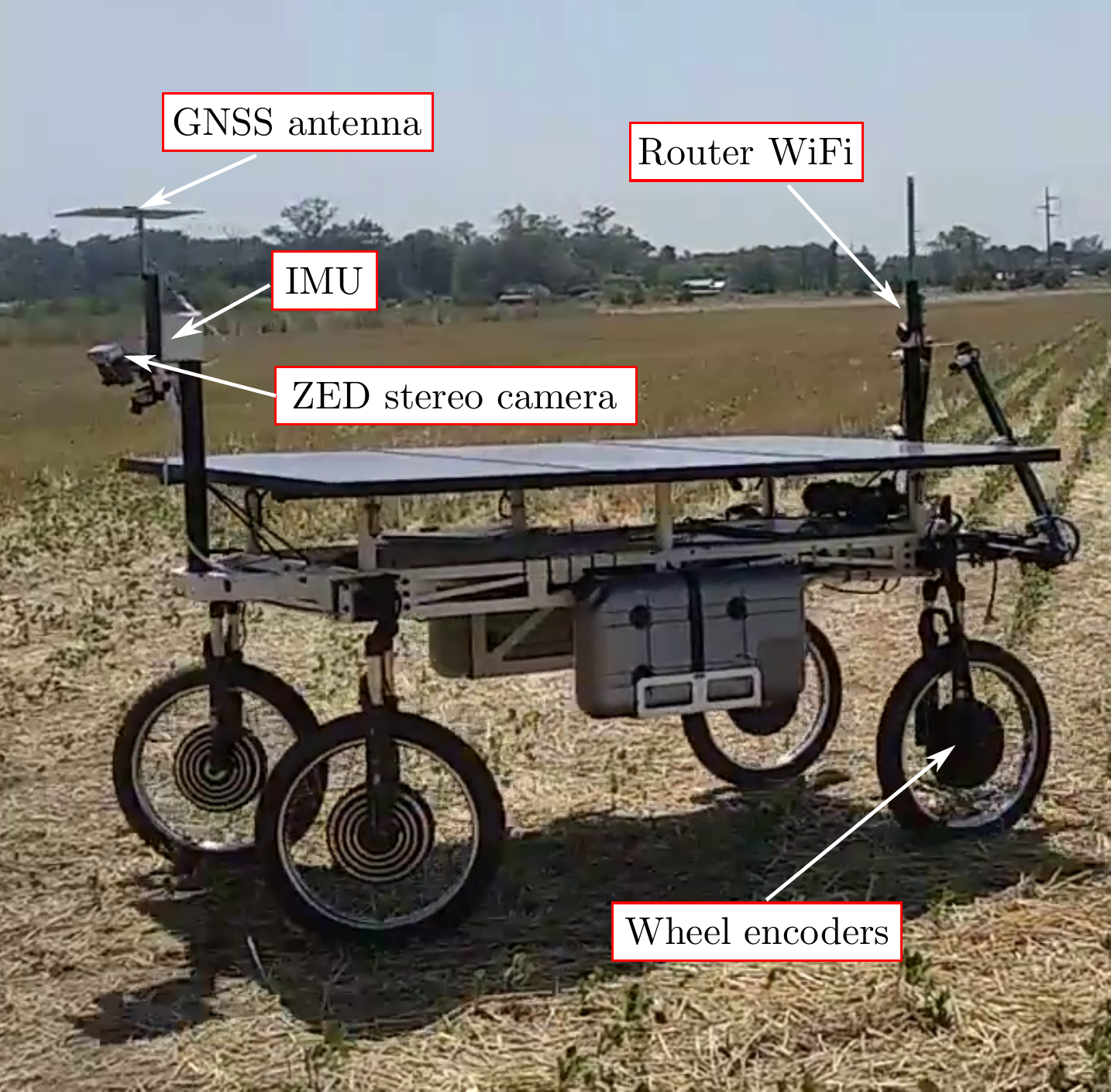}
    \caption{Weed removal robot used in our in-house dataset in a soybean field.}
    \label{fig:robot}
\end{figure}

\begin{figure}[t]
    \centering
    \includegraphics[width=\columnwidth]{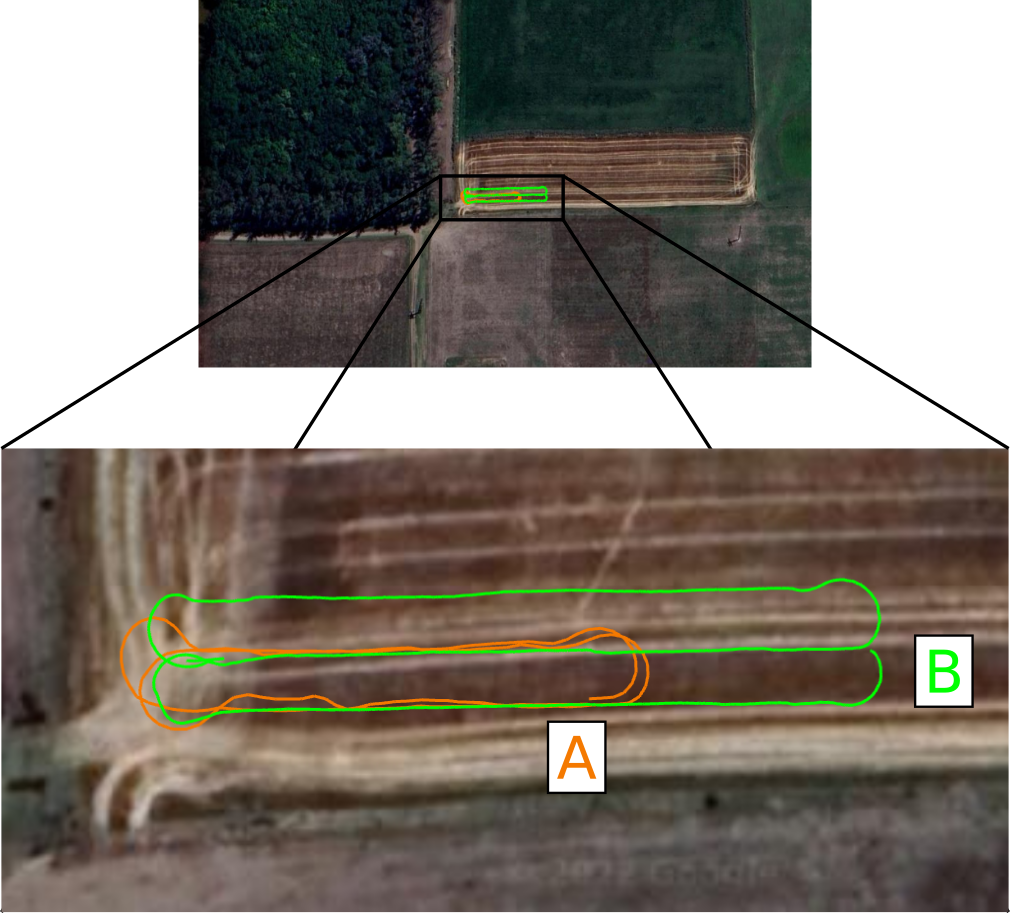}
    \caption{GNSS-RTK trajectories for the sequences A (orange) and B (green) of the in-house recordings in the soybean field.}
    \label{fig:satellital}
\end{figure}

\begin{figure}[tp]
    \centering
    \subfloat{\includegraphics[width=0.45\columnwidth]{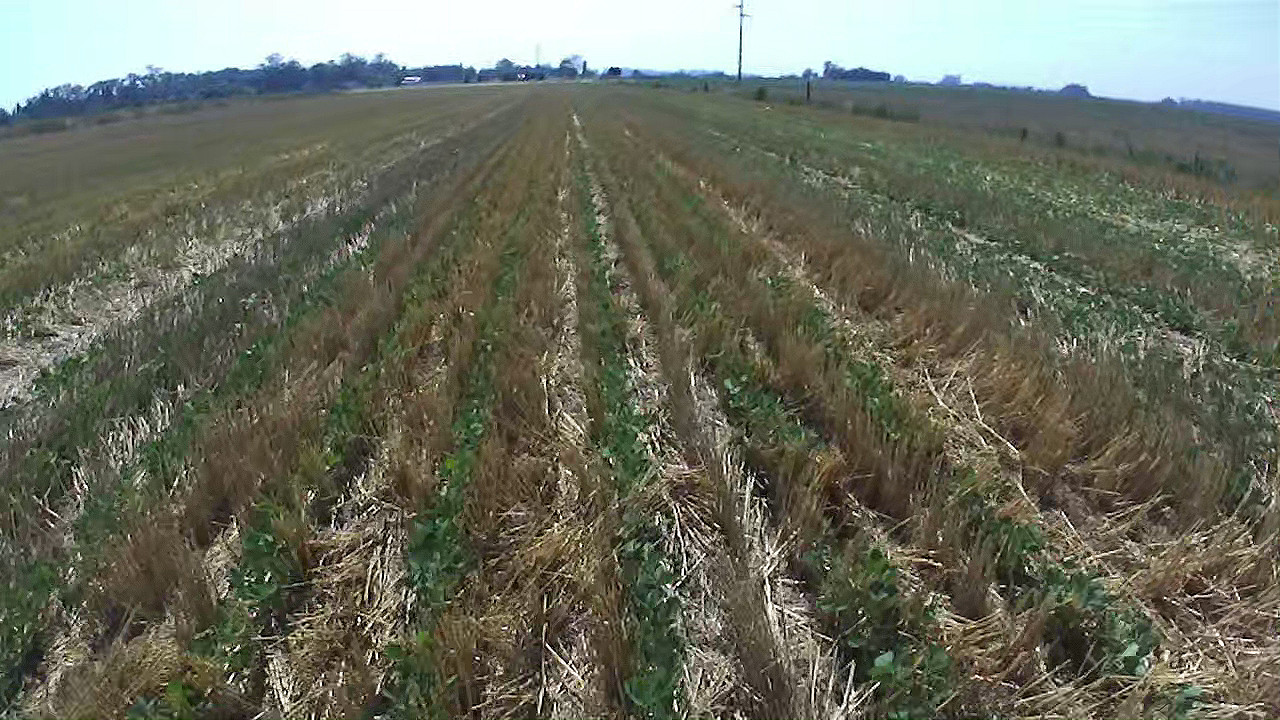}}
    \hspace{0.1em}
    \subfloat{\includegraphics[width=0.45\columnwidth]{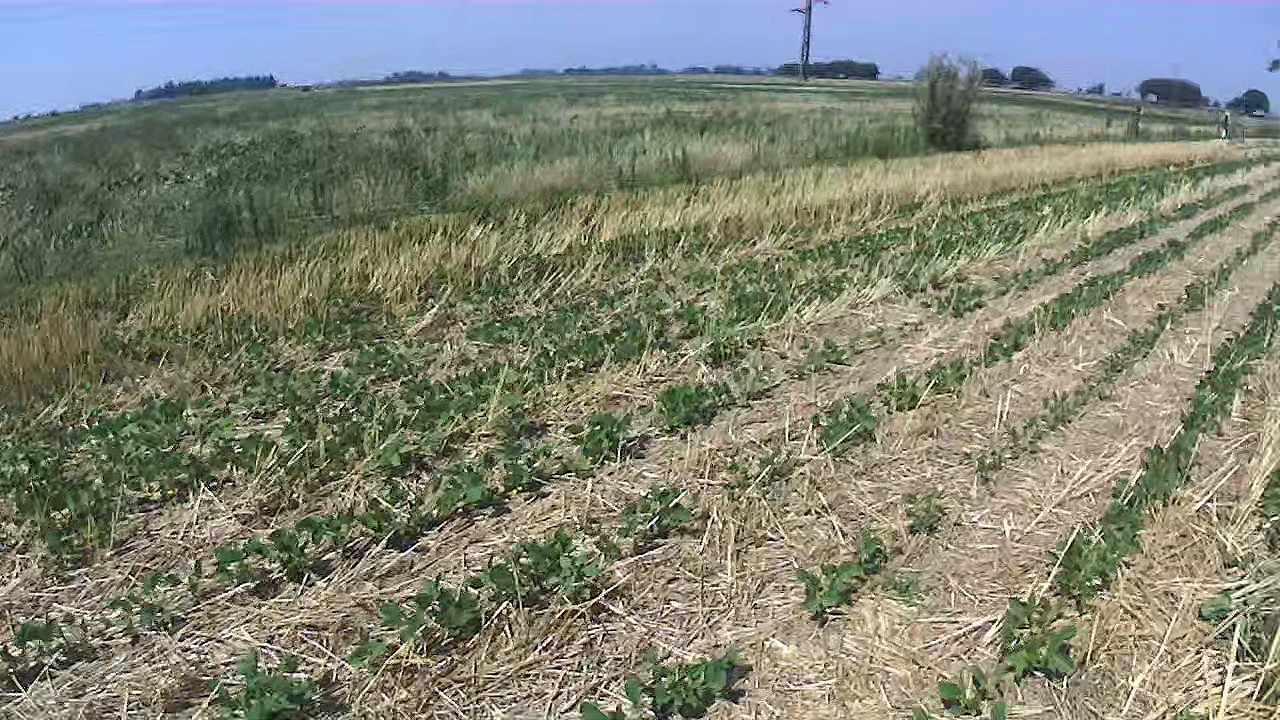}}\\
    \subfloat{\includegraphics[width=0.45\columnwidth]{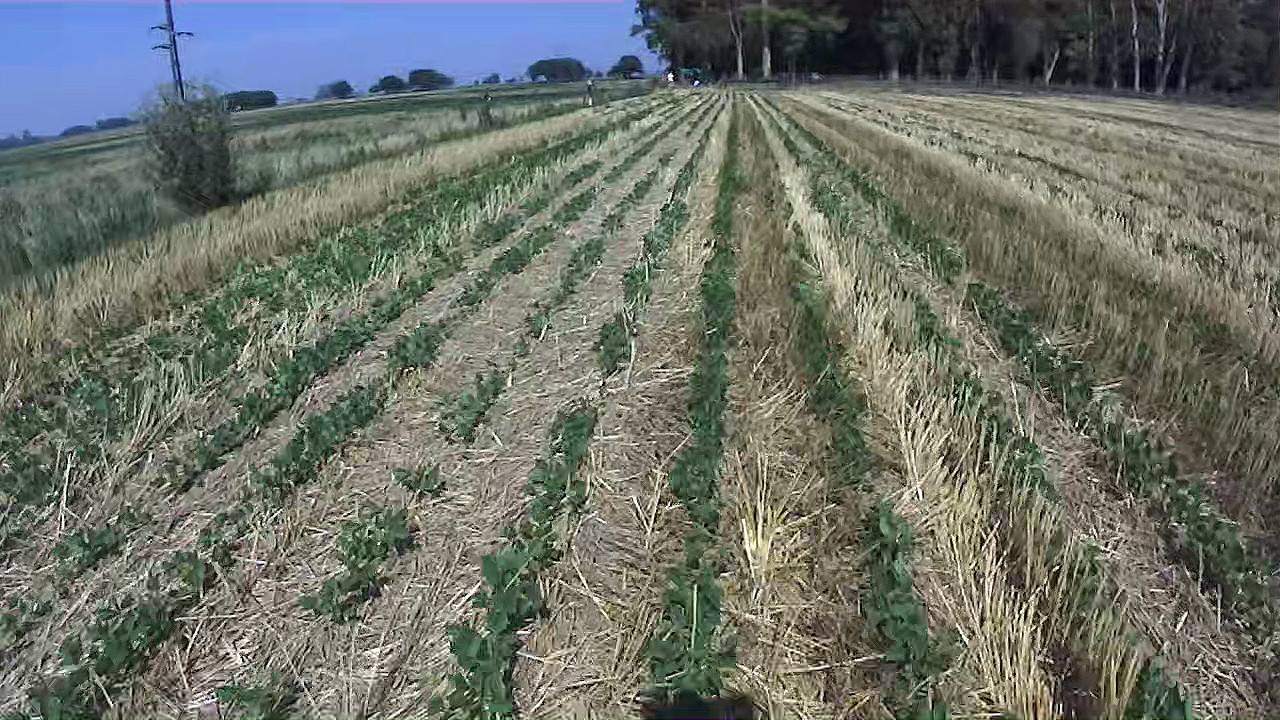}}
    \hspace{0.1em}
    \subfloat{\includegraphics[width=0.45\columnwidth]{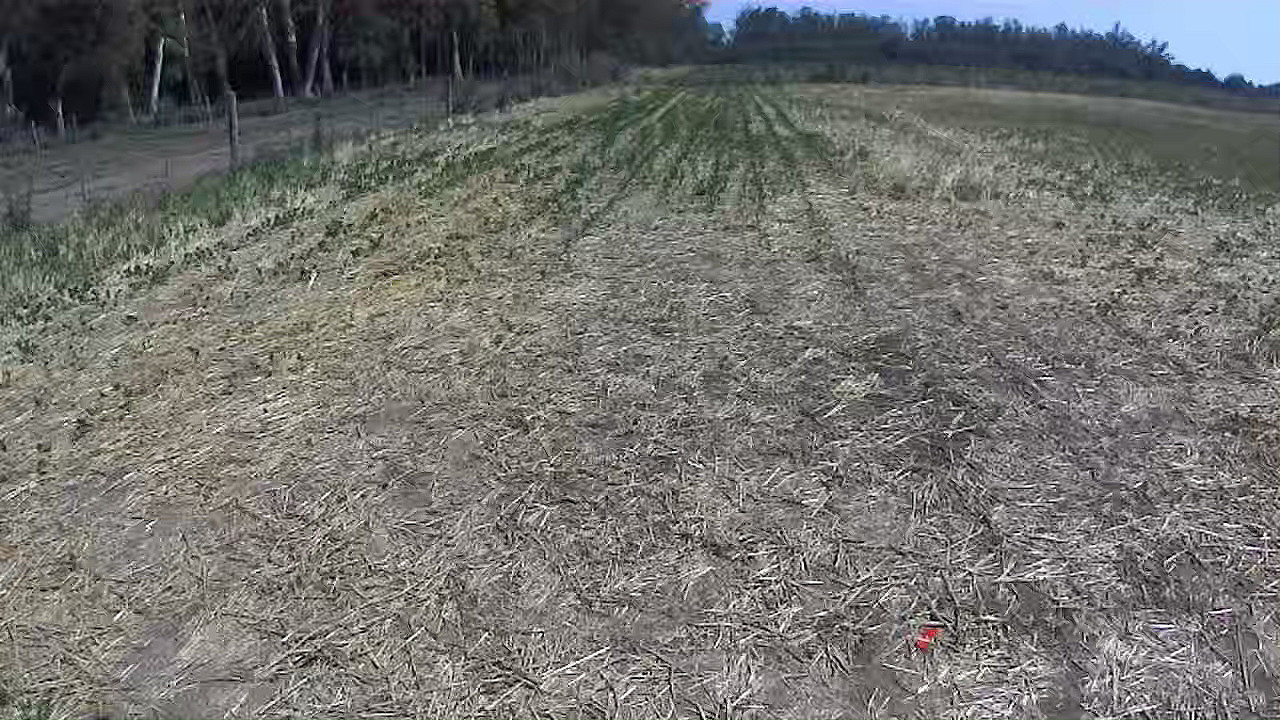}}\\
    \subfloat{\includegraphics[width=0.45\columnwidth]{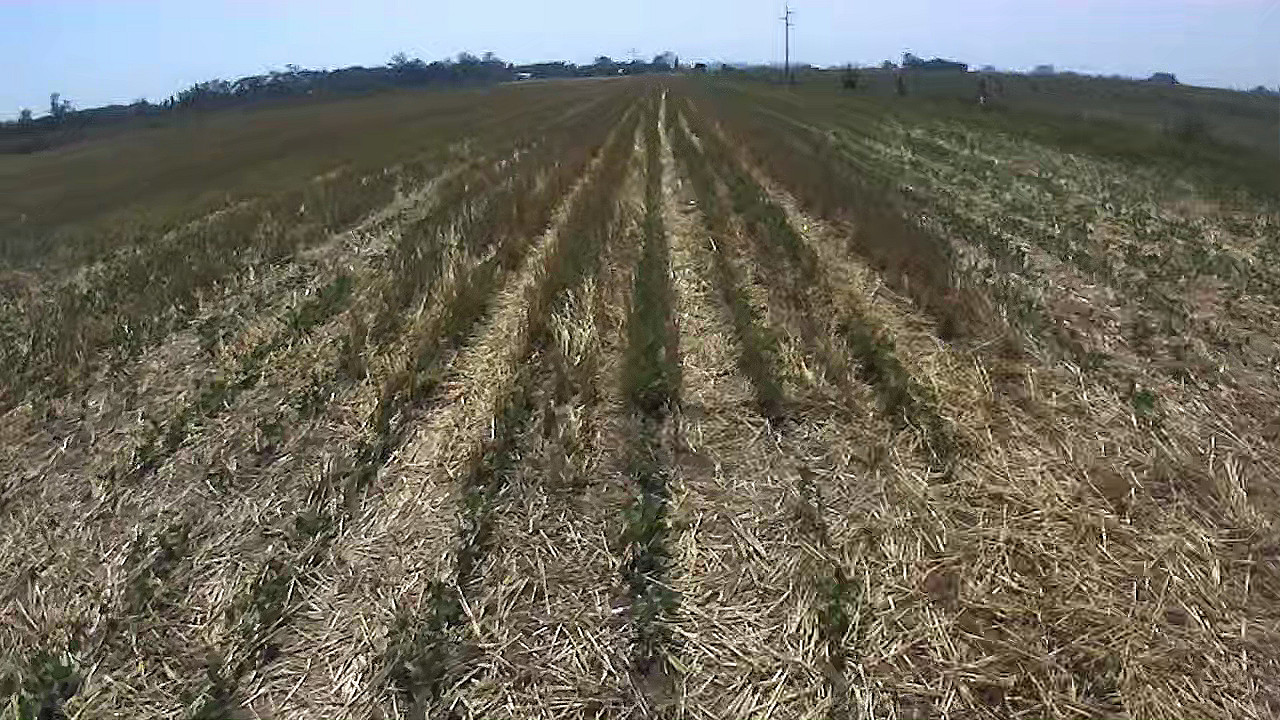}}
    \hspace{0.1em}
    \subfloat{\includegraphics[width=0.45\columnwidth]{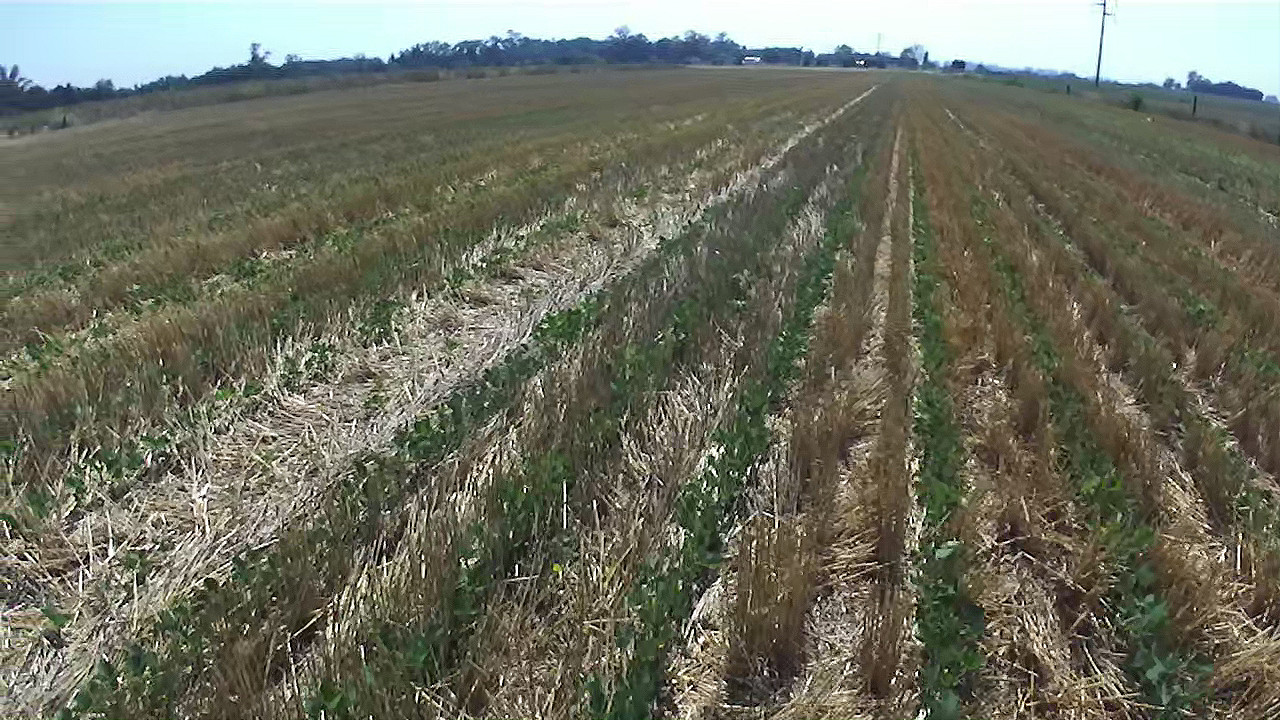}}
    \caption{Sample images from our in-house dataset. Note the repetitive textures, a challenge for visual SLAM.}
    \label{fig:image_samples}
\end{figure}

In the experiments from the previous section, noisy GNSS measurements had to be simulated from GNSS-RTK ones, as the dataset does not contain conventional GNSS measurements. In this section we present an evaluation with conventional GNSS measurements. For this, we equipped our weed removal robot with such sensor and deployed it again in a soybean field. On board the robot there is a ZED stereo camera which captures images $\si{1280}\times\SI{720}{\px}$ at \SI{15}{\hertz}, an Emlid Reach GNSS operating at a frequency of \SI{5}{\hertz}, and an InvenSense MPU-9250 IMU set at \SI{200}{\hertz}. The covariance of the conventional GNSS measurements is offered by the driver of the GNSS receiver. In addition, there is an GNSS-RTK used as positional ground-truth. Figure~\ref{fig:robot} shows the robot configuration in the soybean field. We commanded the robot to record two data sequences. The corresponding GNSS-RTK trajectories are shown in Figure~\ref{fig:satellital} and images samples captured by the ZED camera can be seen in Figure~\ref{fig:image_samples}.

\begin{figure*}[!btp]
    \centering
    \subfloat[Sequence A\label{trajectories_sequence_A}]{\includegraphics[width=0.42\textwidth]{images/zavalla_a.pdf}}
  \hspace{1cm}
  \subfloat[Sequence B\label{trajectories_sequence_B}]{\includegraphics[width=0.42\textwidth]{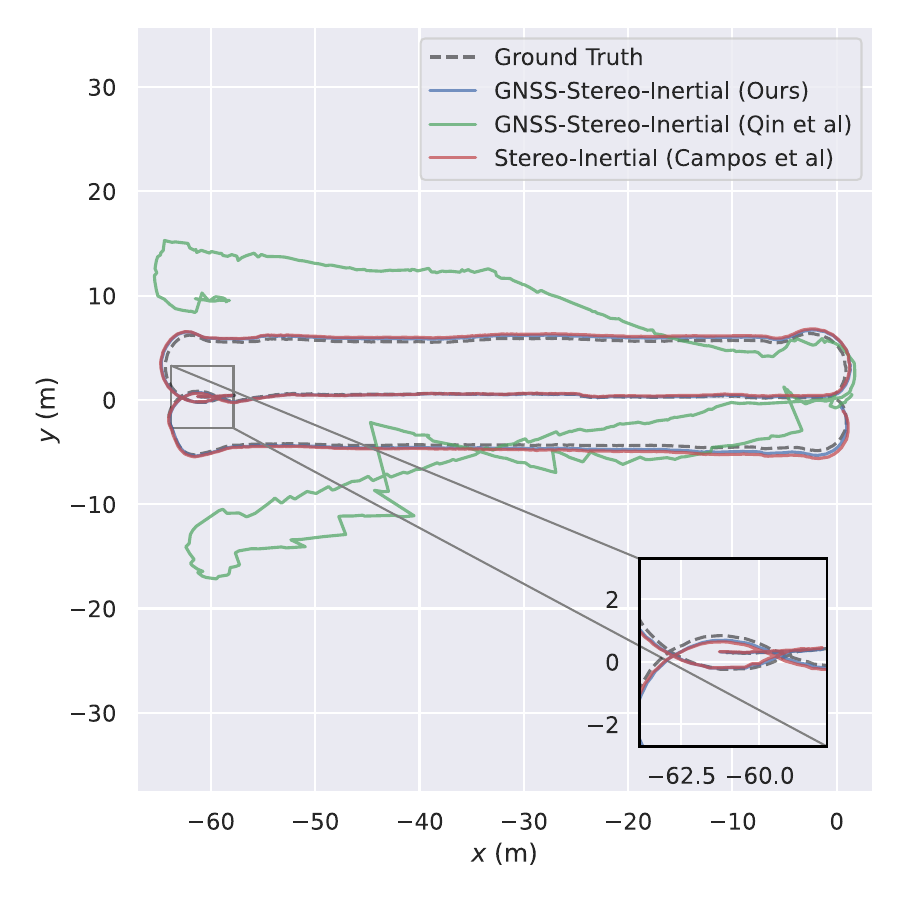}}\\
    \caption{Results from Stereo-Inertial ORB-SLAM3 \cite{campos2021orbslam3}, the loosely-coupled GNSS-Stereo-Inertial system of \cite{qin2019general} and our tightly-coupled GNSS-Stereo-Inertial implementation on our in-house recordings in soybean fields, using conventional GNSS. Note the smaller errors of tightly-coupled approaches, and how our GNSS fusion improves over the stereo-inertial baseline.}
    \label{fig:trajectories_zavalla}
\end{figure*}

On this data we ran the three frameworks mentioned in the previous experiment. The results of this experiment are shown in Table~\ref{tab:real_gps_results}, while the trajectories can be seen in the Figure~\ref{fig:trajectories_zavalla}. Estimated trajectories were aligned again with the ground-truth using
Umeyama's method.
\begin{table}[!tp]
    \centering
    \caption{Mean and standard deviation (between parentheses) of the Absolute Trajectory Error (ATE) [\si{\metre}] for stereo-inertial ORB-SLAM3 \cite{campos2021orbslam3}, a loosely-coupled GNSS-stereo-inertial system \cite{qin2019general} and our tightly-coupled GNSS-stereo-inertial framework in the in-house recordings in soybean fields. Best results are in \bf{bold}.}
    
    \resizebox{\linewidth}{!} {
        \begin{tabular}{cccc}
        \hline
                    Sequence & Stereo-Inertial & GNSS-Stereo-Inertial & GNSS-Stereo-Inertial \\
                    & \cite{campos2021orbslam3} & \cite{qin2019general} & (Ours) \\ \hline
        A &  0.64 (0.33) &  1.08 (0.78)                &  \bf{0.44 (0.16)} \\
        B & 0.43 (0.18)  &    5.58 (3.57)              & \bf{0.36 (0.13)} 
        \\
 \hline
        \end{tabular}
    }
    \label{tab:real_gps_results}
\end{table}

\subsection{Discussion}
As can be seen in the results, our implementation clearly outperforms the stereo-inertial configuration of ORB-SLAM3 and the loosely-coupled approach in \cite{qin2019general}. As a very relevant note, we ran the full stereo-inertial ORB-SLAM3 in our configuration sequences with loop closure capabilities and, with its configuration by default, it was unable to detect previously visited locations and hence close loops due to insufficiently discriminative visual appearances of the agricultural environment (\emph{perceptual aliasing}). Although the default configuration for the loop closure parameters might be loosened to detect a higher number of loop closures, that would also produce a higher number of false positives (due again to perceptual aliasing) that would corrupt the estimation. These challenges are the main motivation for incorporating global positioning sensors in agricultural environments, allowing to reduce the drift without depending on visual features. Very interestingly, we found in our experiments that only one third of the optimized keyframes had associated GNSS measurements. This may indicate that high-frequency GNSS measurements are not necessary to improve the estimation of visual-inertial SLAM, and a sparse subset of them might suffice to offer a reasonable performance.

Unlike the loosely-coupled system, our implementation returns smoother trajectories. Moreover, since the fusion is loosely-coupled, the global position measurements correct the estimate without considering the continuous motion of the robot and acts as an interpolation between the underlying visual-inertial system and the GNSS measurement. Even though in sequence 02 of the Rosario Dataset, the loosely-coupled fusion system obtains a lower error, in the trajectory of the Figure~\ref{fig:trajectories_rosario} it can be observed that the estimation looks bumpy. Smooth pose estimation, like the one offered by our tightly-coupled approach, is more suitable for use in a navigation control algorithm.

Regarding the experiment with conventional GNSS measurements, it should be pointed out that the loosely-coupled system lost the visual-inertial tracking in the two sequences. This indicates that it is important not only to focus on global measurements, but also to have a robust visual-inertial fusion. In our case, we use ORB-SLAM3 as the underlying system, as a result of having analyzed the performance of different visual-inertial systems in previous research \cite{cremona2022evaluation}. As a conclusion, in addition to a tight coupling of the sensor data, the robustness of the visual-inertial estimates are also relevant for practical implementations in agricultural applications.

An important consideration is the modelling of the noise of GNSS measurements. Based on previous works \cite{cioffi2020tightly, boche2022dropout}, the uncertainty was modelled as additive isotropic Gaussian noise. This is a simple model that arises naturally from the GNSS device data, as the device drivers generally provide a covariance of the position. Other ways of modelling the noise of GNSS measurements in the context of pose estimation are worth studying, as when comparing the simulated signal in the section \ref{sec:experiments_rosario_dataset} experiment with the conventional GNSS signal used in the field experiments, differences in their behaviour were observed. When the conventional GNSS signal was inspected in detail, a bias was found, mainly at altitude, which could be verified by the GNSS-RTK. Therefore, this topic should be addressed in future work.

\section{Conclusions}
\label{sec:conclusions}
This work presents a GNSS-stereo-inertial SLAM framework that fuses in a tightly-coupled manner the information from a stereo camera, an IMU and a conventional GNSS sensor. In order to report the most competitive results, we implement our GNSS factor on top of the ORB-SLAM3 framework, the top performer in the evaluation of \cite{cremona2022evaluation}. As we are motivated by long-term autonomous navigation in arable farms, we present results in the Rosario Dataset and in-house sequences from an agricultural robot. Very importantly, several works in the literature evaluate GNSS-stereo-inertial SLAM methods by emulating conventional GNSS measurements while we use a real sensor, so we are the first ones in reporting results in realistic conditions in agricultural scenes.

Our results show that there is a consistent gain in accuracy if GNSS measurements are tightly fused with visual and inertial ones in the local mapping optimization of a SLAM system. Very importantly, not only the localization errors are reduced but also their variance between runs, indicating a looser dependence from the visual features used.

As an additional contribution of this work, we release our implementation for the benefit of the agricultural robotics community.



\section*{Acknowledgments}
\label{sec:acknowledgments}
This work was partially supported by CONICET (Argentina) (PUE 0015-2016), by the Santa Fe province (Argentina) Government under Grant PEICID-2021-170, by the Spanish Government under Grants PGC2018-096367-B-I00 and
PID2021-127685NB-I00 and by the Aragon Government under Grant DGA T45 17R/FSE.

\bibliography{biblio}

\begin{thebibliography}{29}
\expandafter\ifx\csname natexlab\endcsname\relax\def\natexlab#1{#1}\fi
\providecommand{\url}[1]{\texttt{#1}}
\providecommand{\href}[2]{#2}
\providecommand{\path}[1]{#1}
\providecommand{\DOIprefix}{doi:}
\providecommand{\ArXivprefix}{arXiv:}
\providecommand{\URLprefix}{URL: }
\providecommand{\Pubmedprefix}{pmid:}
\providecommand{\doi}[1]{\href{http://dx.doi.org/#1}{\path{#1}}}
\providecommand{\Pubmed}[1]{\href{pmid:#1}{\path{#1}}}
\providecommand{\bibinfo}[2]{#2}
\ifx\xfnm\relax \def\xfnm[#1]{\unskip,\space#1}\fi
\bibitem[{{Auat Cheein} \& {Carelli}(2013)}]{carelli2013agriculturalrobotics}
\bibinfo{author}{{Auat Cheein}, F.~A.}, \& \bibinfo{author}{{Carelli}, R.}
  (\bibinfo{year}{2013}).
\newblock
  \bibinfo{title}{\href{https://doi.org/10.1109/MIE.2013.2252957}{Agricultural
  Robotics: Unmanned Robotic Service Units in Agricultural Tasks}}.
\newblock {\it \bibinfo{journal}{{IEEE} Industrial Electronics Magazine}\/},
  {\it \bibinfo{volume}{7}\/}, \bibinfo{pages}{48--58}.
  \DOIprefix\doi{10.1109/MIE.2013.2252957}.
\bibitem[{Bac et~al.(2014)Bac, van Henten, Hemming \&
  Edan}]{wouter2014harvesting}
\bibinfo{author}{Bac, C.~W.}, \bibinfo{author}{van Henten, E.~J.},
  \bibinfo{author}{Hemming, J.}, \& \bibinfo{author}{Edan, Y.}
  (\bibinfo{year}{2014}).
\newblock \bibinfo{title}{\href{https://doi.org/10.1002/rob.21525}{Harvesting
  Robots for High-value Crops: State-of-the-art Review and Challenges Ahead}}.
\newblock {\it \bibinfo{journal}{Journal of Field Robotics}\/},  {\it
  \bibinfo{volume}{31}\/}, \bibinfo{pages}{888--911}.
  \DOIprefix\doi{10.1002/rob.21525}.
\bibitem[{Boche et~al.(2022)Boche, Zuo, Schaefer \&
  Leutenegger}]{boche2022dropout}
\bibinfo{author}{Boche, S.}, \bibinfo{author}{Zuo, X.},
  \bibinfo{author}{Schaefer, S.}, \& \bibinfo{author}{Leutenegger, S.}
  (\bibinfo{year}{2022}).
\newblock
  \bibinfo{title}{\href{https://doi.org/10.48550/arxiv.2208.00709}{Visual-Inertial
  SLAM with Tightly-Coupled Dropout-Tolerant GPS Fusion}}.
\newblock \URLprefix \url{https://arxiv.org/abs/2208.00709}.
  \DOIprefix\doi{10.48550/ARXIV.2208.00709}.
\bibitem[{Cadena et~al.(2016)Cadena, Carlone, Carrillo, Latif, Scaramuzza,
  Neira, Reid \& Leonard}]{cadena2016past}
\bibinfo{author}{Cadena, C.}, \bibinfo{author}{Carlone, L.},
  \bibinfo{author}{Carrillo, H.}, \bibinfo{author}{Latif, Y.},
  \bibinfo{author}{Scaramuzza, D.}, \bibinfo{author}{Neira, J.},
  \bibinfo{author}{Reid, I.}, \& \bibinfo{author}{Leonard, J.~J.}
  (\bibinfo{year}{2016}).
\newblock \bibinfo{title}{\href{https://doi.org/10.1109/TRO.2016.2624754}{Past,
  Present, and Future of Simultaneous Localization and Mapping: Toward the
  Robust-Perception Age}}.
\newblock {\it \bibinfo{journal}{{IEEE} Trans. Robotics}\/},  {\it
  \bibinfo{volume}{32}\/}, \bibinfo{pages}{1309--1332}.
  \DOIprefix\doi{10.1109/TRO.2016.2624754}.
\bibitem[{Campos et~al.(2021)Campos, Elvira, Rodríguez, M.~Montiel \&
  D.~Tardós}]{campos2021orbslam3}
\bibinfo{author}{Campos, C.}, \bibinfo{author}{Elvira, R.},
  \bibinfo{author}{Rodríguez, J. J.~G.}, \bibinfo{author}{M.~Montiel, J.~M.},
  \& \bibinfo{author}{D.~Tardós, J.} (\bibinfo{year}{2021}).
\newblock
  \bibinfo{title}{\href{https://doi.org/10.1109/TRO.2021.3075644}{ORB-SLAM3: An
  Accurate Open-Source Library for Visual, Visual–Inertial, and Multimap
  SLAM}}.
\newblock {\it \bibinfo{journal}{{IEEE} Trans. Robotics}\/},  {\it
  \bibinfo{volume}{37}\/}, \bibinfo{pages}{1874--1890}.
  \DOIprefix\doi{10.1109/TRO.2021.3075644}.
\bibitem[{Cao et~al.(2022)Cao, Lu \& Shen}]{cao2021gvins}
\bibinfo{author}{Cao, S.}, \bibinfo{author}{Lu, X.}, \& \bibinfo{author}{Shen,
  S.} (\bibinfo{year}{2022}).
\newblock
  \bibinfo{title}{\href{https://doi.org/10.1109/TRO.2021.3133730}{GVINS:
  Tightly Coupled GNSS-Visual-Inertial Fusion for Smooth and Consistent State
  Estimation}}.
\newblock {\it \bibinfo{journal}{IEEE Transactions on Robotics}\/},  {\it
  \bibinfo{volume}{38}\/}, \bibinfo{pages}{2004--2021}.
  \DOIprefix\doi{10.1109/TRO.2021.3133730}.
\bibitem[{Cioffi \& Scaramuzza(2020)}]{cioffi2020tightly}
\bibinfo{author}{Cioffi, G.}, \& \bibinfo{author}{Scaramuzza, D.}
  (\bibinfo{year}{2020}).
\newblock
  \bibinfo{title}{\href{https://doi.org/10.1109/IROS45743.2020.9341697}{Tightly-coupled
  Fusion of Global Positional Measurements in Optimization-based
  Visual-Inertial Odometry}}.
\newblock In {\it \bibinfo{booktitle}{IEEE/RSJ Intl. Conf. on Intelligent
  Robots and Systems (IROS)}\/} (pp. \bibinfo{pages}{5089--5095}).
\newblock \DOIprefix\doi{10.1109/IROS45743.2020.9341697}.
\bibitem[{Cremona et~al.(2022)Cremona, Comelli \& Pire}]{cremona2022evaluation}
\bibinfo{author}{Cremona, J.}, \bibinfo{author}{Comelli, R.}, \&
  \bibinfo{author}{Pire, T.} (\bibinfo{year}{2022}).
\newblock
  \bibinfo{title}{\href{{https://doi.org/10.1002/rob.22099}}{Experimental
  evaluation of Visual-Inertial Odometry systems for arable farming}}.
\newblock {\it \bibinfo{journal}{Journal of Field Robotics}\/},  {\it
  \bibinfo{volume}{39}\/}, \bibinfo{pages}{1123--1137}. \URLprefix
  \url{https://onlinelibrary.wiley.com/doi/abs/10.1002/rob.22099}.
  \DOIprefix\doi{10.1002/rob.22099}.
\bibitem[{{Forster} et~al.(2017){Forster}, {Carlone}, {Dellaert} \&
  {Scaramuzza}}]{forster2017onmanifold}
\bibinfo{author}{{Forster}, C.}, \bibinfo{author}{{Carlone}, L.},
  \bibinfo{author}{{Dellaert}, F.}, \& \bibinfo{author}{{Scaramuzza}, D.}
  (\bibinfo{year}{2017}).
\newblock
  \bibinfo{title}{\href{https://doi.org/10.1109/TRO.2016.2597321}{On-Manifold
  Preintegration for Real-Time Visual--Inertial Odometry}}.
\newblock {\it \bibinfo{journal}{{IEEE} Trans. Robotics}\/},  {\it
  \bibinfo{volume}{33}\/}, \bibinfo{pages}{1--21}.
  \DOIprefix\doi{10.1109/TRO.2016.2597321}.
\bibitem[{Lee et~al.(2020)Lee, Eckenhoff, Geneva \&
  Huang}]{lee2020intermittent}
\bibinfo{author}{Lee, W.}, \bibinfo{author}{Eckenhoff, K.},
  \bibinfo{author}{Geneva, P.}, \& \bibinfo{author}{Huang, G.}
  (\bibinfo{year}{2020}).
\newblock
  \bibinfo{title}{\href{https://doi.org/10.1109/ICRA40945.2020.9197029}{Intermittent
  GPS-aided VIO: Online Initialization and Calibration}}.
\newblock In {\it \bibinfo{booktitle}{IEEE Intl. Conf. on Robotics and
  Automation (ICRA)}\/} (pp. \bibinfo{pages}{5724--5731}).
\newblock \DOIprefix\doi{10.1109/ICRA40945.2020.9197029}.
\bibitem[{Leutenegger(2022)}]{leutenegger2022okvis2}
\bibinfo{author}{Leutenegger, S.} (\bibinfo{year}{2022}).
\newblock
  \bibinfo{title}{\href{https://doi.org/10.48550/arxiv.2202.09199}{OKVIS2:
  Realtime Scalable Visual-Inertial SLAM with Loop Closure}}.
\newblock \URLprefix \url{https://arxiv.org/abs/2202.09199}.
  \DOIprefix\doi{10.48550/ARXIV.2202.09199}.
\bibitem[{Li et~al.(2019)Li, Zhang, Gao, Niu \& El-sheimy}]{li2019tight}
\bibinfo{author}{Li, T.}, \bibinfo{author}{Zhang, H.}, \bibinfo{author}{Gao,
  Z.}, \bibinfo{author}{Niu, X.}, \& \bibinfo{author}{El-sheimy, N.}
  (\bibinfo{year}{2019}).
\newblock \bibinfo{title}{\href{https://doi.org/10.3390/rs11060610}{Tight
  Fusion of a Monocular Camera, MEMS-IMU, and Single-Frequency Multi-GNSS RTK
  for Precise Navigation in GNSS-Challenged Environments}}.
\newblock {\it \bibinfo{journal}{Remote Sensing}\/},  {\it
  \bibinfo{volume}{11}\/}. \DOIprefix\doi{10.3390/rs11060610}.
\bibitem[{Liu et~al.(2021)Liu, Gao \& Hu}]{liu2021optimization}
\bibinfo{author}{Liu, J.}, \bibinfo{author}{Gao, W.}, \& \bibinfo{author}{Hu,
  Z.} (\bibinfo{year}{2021}).
\newblock
  \bibinfo{title}{\href{https://doi.org/10.1109/ICRA48506.2021.9562013}{Optimization-Based
  Visual-Inertial SLAM Tightly Coupled with Raw GNSS Measurements}}.
\newblock In {\it \bibinfo{booktitle}{IEEE Intl. Conf. on Robotics and
  Automation (ICRA)}\/} (pp. \bibinfo{pages}{11612--11618}).
\newblock \DOIprefix\doi{10.1109/ICRA48506.2021.9562013}.
\bibitem[{Lynen et~al.(2013)Lynen, Achtelik, Weiss, Chli \&
  Siegwart}]{lynen2013robust}
\bibinfo{author}{Lynen, S.}, \bibinfo{author}{Achtelik, M.~W.},
  \bibinfo{author}{Weiss, S.}, \bibinfo{author}{Chli, M.}, \&
  \bibinfo{author}{Siegwart, R.} (\bibinfo{year}{2013}).
\newblock \bibinfo{title}{\href{https://doi.org/10.1109/IROS.2013.6696917}{A
  robust and modular multi-sensor fusion approach applied to MAV navigation}}.
\newblock In {\it \bibinfo{booktitle}{IEEE/RSJ Intl. Conf. on Intelligent
  Robots and Systems (IROS)}\/} (pp. \bibinfo{pages}{3923--3929}).
\newblock \DOIprefix\doi{10.1109/IROS.2013.6696917}.
\bibitem[{Mascaro et~al.(2018)Mascaro, Teixeira, Hinzmann, Siegwart \&
  Chli}]{mascaro2018gomsf}
\bibinfo{author}{Mascaro, R.}, \bibinfo{author}{Teixeira, L.},
  \bibinfo{author}{Hinzmann, T.}, \bibinfo{author}{Siegwart, R.}, \&
  \bibinfo{author}{Chli, M.} (\bibinfo{year}{2018}).
\newblock
  \bibinfo{title}{\href{https://doi.org/10.1109/ICRA.2018.8460193}{GOMSF:
  Graph-Optimization Based Multi-Sensor Fusion for robust UAV Pose
  estimation}}.
\newblock In {\it \bibinfo{booktitle}{IEEE Intl. Conf. on Robotics and
  Automation (ICRA)}\/} (pp. \bibinfo{pages}{1421--1428}).
\newblock \DOIprefix\doi{10.1109/ICRA.2018.8460193}.
\bibitem[{{Mur-Artal} \& {Tardós}(2017{\natexlab{a}})}]{mur2017visual}
\bibinfo{author}{{Mur-Artal}, R.}, \& \bibinfo{author}{{Tardós}, J.~D.}
  (\bibinfo{year}{2017}{\natexlab{a}}).
\newblock
  \bibinfo{title}{\href{https://doi.org/10.1109/LRA.2017.2653359}{Visual-Inertial
  Monocular SLAM With Map Reuse}}.
\newblock {\it \bibinfo{journal}{(IEEE) Robotics and Automation Letters}\/},
  {\it \bibinfo{volume}{2}\/}, \bibinfo{pages}{796--803}.
  \DOIprefix\doi{10.1109/LRA.2017.2653359}.
\bibitem[{{Mur-Artal} \& {Tardós}(2017{\natexlab{b}})}]{mur2017orb}
\bibinfo{author}{{Mur-Artal}, R.}, \& \bibinfo{author}{{Tardós}, J.~D.}
  (\bibinfo{year}{2017}{\natexlab{b}}).
\newblock
  \bibinfo{title}{\href{https://doi.org/10.1109/TRO.2017.2705103}{ORB-SLAM2: An
  Open-Source SLAM System for Monocular, Stereo, and RGB-D Cameras}}.
\newblock {\it \bibinfo{journal}{{IEEE} Trans. Robotics}\/},  {\it
  \bibinfo{volume}{33}\/}, \bibinfo{pages}{1255--1262}.
  \DOIprefix\doi{10.1109/TRO.2017.2705103}.
\bibitem[{Pire et~al.(2019)Pire, Mujica, Civera \& Kofman}]{pire2019rosario}
\bibinfo{author}{Pire, T.}, \bibinfo{author}{Mujica, M.},
  \bibinfo{author}{Civera, J.}, \& \bibinfo{author}{Kofman, E.}
  (\bibinfo{year}{2019}).
\newblock \bibinfo{title}{\href{https://doi.org/10.1177/0278364919841437}{The
  Rosario Dataset: Multisensor Data for Localization and Mapping in
  Agricultural Environments}}.
\newblock {\it \bibinfo{journal}{Intl. J. of Robotics Research}\/},  {\it
  \bibinfo{volume}{38}\/}, \bibinfo{pages}{633--641}.
  \DOIprefix\doi{10.1177/0278364919841437}.
\bibitem[{Qin et~al.(2019)Qin, Cao, Pan \& Shen}]{qin2019general}
\bibinfo{author}{Qin, T.}, \bibinfo{author}{Cao, S.}, \bibinfo{author}{Pan,
  J.}, \& \bibinfo{author}{Shen, S.} (\bibinfo{year}{2019}).
\newblock \bibinfo{title}{\href{https://arxiv.org/abs/1901.03642}{A General
  Optimization-based Framework for Global Pose Estimation with Multiple
  Sensors}}.
\newblock \href{http://arxiv.org/abs/1901.03642}{\tt arXiv:1901.03642}.
\bibitem[{{Qin} et~al.(2018){Qin}, {Li} \& {Shen}}]{qin2018vins}
\bibinfo{author}{{Qin}, T.}, \bibinfo{author}{{Li}, P.}, \&
  \bibinfo{author}{{Shen}, S.} (\bibinfo{year}{2018}).
\newblock
  \bibinfo{title}{\href{https://doi.org/10.1109/TRO.2018.2853729}{VINS-Mono: A
  Robust and Versatile Monocular Visual-Inertial State Estimator}}.
\newblock {\it \bibinfo{journal}{{IEEE} Trans. Robotics}\/},  {\it
  \bibinfo{volume}{34}\/}, \bibinfo{pages}{1004--1020}.
  \DOIprefix\doi{10.1109/TRO.2018.2853729}.
\bibitem[{Salehi et~al.(2017)Salehi, Gay-bellile, Bourgeois \&
  Chausse}]{salehi2017hybrid}
\bibinfo{author}{Salehi, A.}, \bibinfo{author}{Gay-bellile, V.},
  \bibinfo{author}{Bourgeois, S.}, \& \bibinfo{author}{Chausse, F.}
  (\bibinfo{year}{2017}).
\newblock \bibinfo{title}{\href{https://doi.org/10.1109/IVS.2017.7995957}{A
  hybrid bundle adjustment/pose-graph approach to VSLAM/GPS fusion for
  low-capacity platforms}}.
\newblock In {\it \bibinfo{booktitle}{IEEE Intelligent Vehicles Symposium
  (IV)}\/} (pp. \bibinfo{pages}{1728--1735}).
\newblock \bibinfo{organization}{IEEE}.
\newblock \DOIprefix\doi{10.1109/IVS.2017.7995957}.
\bibitem[{Shen et~al.(2014)Shen, Mulgaonkar, Michael \& Kumar}]{shen2014multi}
\bibinfo{author}{Shen, S.}, \bibinfo{author}{Mulgaonkar, Y.},
  \bibinfo{author}{Michael, N.}, \& \bibinfo{author}{Kumar, V.}
  (\bibinfo{year}{2014}).
\newblock
  \bibinfo{title}{\href{https://doi.org/10.1109/ICRA.2014.6907588}{Multi-sensor
  fusion for robust autonomous flight in indoor and outdoor environments with a
  rotorcraft MAV}}.
\newblock In {\it \bibinfo{booktitle}{IEEE Intl. Conf. on Robotics and
  Automation (ICRA)}\/} (pp. \bibinfo{pages}{4974--4981}).
\newblock \DOIprefix\doi{10.1109/ICRA.2014.6907588}.
\bibitem[{Strasdat et~al.(2012)Strasdat, Montiel \& Davison}]{strasdat2012why}
\bibinfo{author}{Strasdat, H.}, \bibinfo{author}{Montiel, J.}, \&
  \bibinfo{author}{Davison, A.~J.} (\bibinfo{year}{2012}).
\newblock
  \bibinfo{title}{\href{https://doi.org/10.1016/j.imavis.2012.02.009}{Visual
  SLAM: Why filter?}}
\newblock {\it \bibinfo{journal}{Image and Vision Computing}\/},  {\it
  \bibinfo{volume}{30}\/}, \bibinfo{pages}{65--77}.
  \DOIprefix\doi{10.1016/j.imavis.2012.02.009}.
\bibitem[{{Umeyama}(1991)}]{umeyama1991least}
\bibinfo{author}{{Umeyama}, S.} (\bibinfo{year}{1991}).
\newblock \bibinfo{title}{\href{https://doi.org/10.1109/34.88573}{Least-squares
  estimation of transformation parameters between two point patterns}}.
\newblock {\it \bibinfo{journal}{{IEEE} Trans. Pattern Anal. Machine
  Intell.}\/},  {\it \bibinfo{volume}{13}\/}, \bibinfo{pages}{376--380}.
  \DOIprefix\doi{10.1109/34.88573}.
\bibitem[{Wei et~al.(2011)Wei, Cappelle, Ruichek \& Zann}]{wei2011intelligent}
\bibinfo{author}{Wei, L.}, \bibinfo{author}{Cappelle, C.},
  \bibinfo{author}{Ruichek, Y.}, \& \bibinfo{author}{Zann, F.}
  (\bibinfo{year}{2011}).
\newblock
  \bibinfo{title}{\href{https://doi.org/10.3182/20110828-6-IT-1002.01965}{Intelligent
  Vehicle Localization in Urban Environments Using EKF-based Visual Odometry
  and GPS Fusion}}.
\newblock {\it \bibinfo{journal}{IFAC Proceedings Volumes}\/},  {\it
  \bibinfo{volume}{44}\/}, \bibinfo{pages}{13776--13781}.
  \DOIprefix\doi{10.3182/20110828-6-IT-1002.01965}.
\newblock \bibinfo{note}{18th IFAC World Congress}.
\bibitem[{Weiss et~al.(2012)Weiss, Achtelik, Chli \&
  Siegwart}]{weiss2012versatile}
\bibinfo{author}{Weiss, S.}, \bibinfo{author}{Achtelik, M.~W.},
  \bibinfo{author}{Chli, M.}, \& \bibinfo{author}{Siegwart, R.}
  (\bibinfo{year}{2012}).
\newblock
  \bibinfo{title}{\href{https://doi.org/10.1109/ICRA.2012.6225002}{Versatile
  distributed pose estimation and sensor self-calibration for an autonomous
  MAV}}.
\newblock In {\it \bibinfo{booktitle}{IEEE Intl. Conf. on Robotics and
  Automation (ICRA)}\/} (pp. \bibinfo{pages}{31--38}).
\newblock \DOIprefix\doi{10.1109/ICRA.2012.6225002}.
\bibitem[{Won et~al.(2014{\natexlab{a}})Won, Lee, Heo, Lee, Lee, Kim, Sung \&
  Lee}]{won2014selective}
\bibinfo{author}{Won, D.~H.}, \bibinfo{author}{Lee, E.}, \bibinfo{author}{Heo,
  M.}, \bibinfo{author}{Lee, S.-W.}, \bibinfo{author}{Lee, J.},
  \bibinfo{author}{Kim, J.}, \bibinfo{author}{Sung, S.}, \&
  \bibinfo{author}{Lee, Y.~J.} (\bibinfo{year}{2014}{\natexlab{a}}).
\newblock
  \bibinfo{title}{\href{https://doi.org/10.1109/TIM.2014.2304365}{Selective
  Integration of GNSS, Vision Sensor, and INS Using Weighted DOP Under
  GNSS-Challenged Environments}}.
\newblock {\it \bibinfo{journal}{{IEEE} Transactions on Instrumentation and
  Measurement}\/},  {\it \bibinfo{volume}{63}\/}, \bibinfo{pages}{2288--2298}.
  \DOIprefix\doi{10.1109/TIM.2014.2304365}.
\bibitem[{Won et~al.(2014{\natexlab{b}})Won, Lee, Heo, Sung, Lee \&
  Lee}]{Won2014gnss}
\bibinfo{author}{Won, D.~H.}, \bibinfo{author}{Lee, E.}, \bibinfo{author}{Heo,
  M.}, \bibinfo{author}{Sung, S.}, \bibinfo{author}{Lee, J.}, \&
  \bibinfo{author}{Lee, Y.~J.} (\bibinfo{year}{2014}{\natexlab{b}}).
\newblock \bibinfo{title}{\href{https://doi.org/10.1007/s10291-013-0318-8}{GNSS
  integration with vision-based navigation for low GNSS visibility
  conditions}}.
\newblock {\it \bibinfo{journal}{GPS Solutions}\/},  {\it
  \bibinfo{volume}{18}\/}, \bibinfo{pages}{177--187}. \URLprefix
  \url{https://doi.org/10.1007/s10291-013-0318-8}.
  \DOIprefix\doi{10.1007/s10291-013-0318-8}.
\bibitem[{Yu et~al.(2019)Yu, Gao, Liu, Shen \& Liu}]{yu2019gpsaided}
\bibinfo{author}{Yu, Y.}, \bibinfo{author}{Gao, W.}, \bibinfo{author}{Liu, C.},
  \bibinfo{author}{Shen, S.}, \& \bibinfo{author}{Liu, M.}
  (\bibinfo{year}{2019}).
\newblock
  \bibinfo{title}{\href{https://doi.org/10.1109/IROS40897.2019.8968519}{A
  GPS-aided Omnidirectional Visual-Inertial State Estimator in Ubiquitous
  Environments}}.
\newblock In {\it \bibinfo{booktitle}{IEEE/RSJ Intl. Conf. on Intelligent
  Robots and Systems (IROS)}\/} (pp. \bibinfo{pages}{7750--7755}).
\newblock \DOIprefix\doi{10.1109/IROS40897.2019.8968519}.

\end{thebibliography}
\end{document}